\documentclass[conference,onecolumn]{IEEEtran}
\usepackage[frozencache=true,cachedir=minted-cache]{minted}
\usepackage{amsmath,amssymb,amsfonts}
\usepackage{algorithmic}
\usepackage{graphicx}
\usepackage{textcomp}
\usepackage[dvipsnames,table]{xcolor}
\usepackage{dirtytalk}
\usepackage{circuitikz}
\usepackage{pgfplots}
\usepgfplotslibrary{fillbetween}
\pgfplotsset{width=10cm,compat=1.9}
\usepackage{adjustbox}
\usepackage{siunitx}
\usepackage{fontawesome}
\usepackage{listings}
\usepackage{tabularx}
\usepackage{tabulary}
\usepackage{tabu}
\usepackage{longtable}
\usepackage{tikz}
\usetikzlibrary{shapes.geometric, arrows, positioning}
\usepackage{ragged2e}
\usepackage{circuitikz}
\usepackage{svg}
\usepackage{float}
\usepackage{smartdiagram}
\usepackage[acronym]{glossaries}

\usepackage[backend=biber, style=ieee, url=false]{biblatex}

\newacronym{ABV}{ABV}{Assertion Based Verification}
\newacronym{ADAS}{ADAS}{Advanced Driver Assistance Systems}
\newacronym{ADC}{ADC}{Analog to Digital Converter}
\newacronym{API}{API}{Application Program Interface}
\newacronym{ASIC}{ASIC}{Application-Specific Integrated Circuit}
\newacronym{ASIL}{ASIL}{Automotive Safety Integrity Level}
\newacronym{ATPG}{ATPG}{Automated Test Pattern Generation}
\newacronym{BDD}{BDD}{Binary Decision Diagram}
\newacronym{BMC}{BMC}{Bounded Model Checking}
\newacronym{CPU}{CPU}{Central Processing Unit}
\newacronym{CAD}{CAD}{Computer Aided Design}
\newacronym{CEX}{CEX}{Counter Example}
\newacronym{CNF}{CNF}{Conjunctive Normal Form}
\newacronym{COI}{COI}{Cone of Influence}
\newacronym{ConfigVerMet}{ConfigVerMet}{Configuration Verification Methodology}
\newacronym{CBAW}{CBAW}{Configure-By-Air-Wires}
\newacronym{COV}{COV}{Coverage}
\newacronym{CDV}{CDV}{Coverage Driven Verification}
\newacronym{CSR}{CSR}{Control \& Status Register}
\newacronym{CSV}{CSV}{Comma Separated Values}
\newacronym{CRV}{CRV}{Constrained Random Verification}
\newacronym{CTL}{CTL}{Computational Tree Logic}
\newacronym{DAC}{DAC}{Digital to Analog Converter}
\newacronym{DB}{DB}{Database}
\newacronym{DD}{DD}{Decision Diagram}
\newacronym{DUT}{DUT}{Design Under Test}
\newacronym{DUV}{DUV}{Design Under Verification}
\newacronym{ECC}{ECC}{Error Detection and Correction Code}
\newacronym{EDA}{EDA}{Electronic Design Automation}
\newacronym{FEC}{FEC}{Formal Equivalence Checking}
\newacronym{FEV}{FEV}{Formal Equivalence Verification}
\newacronym{FPV}{FPV}{Formal Property Verification}
\newacronym{FPGA}{FPGA}{Field Programmable Gate Array}
\newacronym{FSM}{FSM}{Finite State Machine}
\newacronym{FuSa}{FuSa}{Functional Safety}
\newacronym{FV}{FV}{Formal Verification}
\newacronym{GUI}{GUI}{Graphical User Interface}
\newacronym{HDL}{HDL}{Hardware Description Language}
\newacronym{HVL}{HVL}{Hardware Verification Language}
\newacronym{IC}{IC}{Integrated Circuit}
\newacronym{IP}{IP}{Intellectual Property}
\newacronym{I2C}{I2C}{Inter-Integrated Circuit}
\newacronym{IEEE}{IEEE}{Institute of Electrical and Electronics Engineers}
\newacronym{ISO}{ISO}{International Organization for Standardization}
\newacronym{ITRS}{ITRS}{International Technology Roadmap for Semiconductors}
\newacronym{JG}{JG}{JasperGold}
\newacronym{LTL}{LTL}{Linear Temporal Logic}
\newacronym{MDV}{MDV}{Metric Driven Verification}
\newacronym{OBDD}{OBDD}{Ordered Binary Decision Diagram}
\newacronym{OVM}{OVM}{Open Verification Methodology}
\newacronym{OVL}{OVL}{Open Verification Library}
\newacronym{PICT}{PICT}{Pairwise Independent Combinatorial Testing}
\newacronym{PSL}{PSL}{Property Specification Language}
\newacronym{QEDTEC}{QEDTEC}{Quad Bit Error Detection, Triple Bit Error Correction}
\newacronym{ROBDD}{ROBDD}{Reduced Ordered Binary Decision Diagram}
\newacronym{RTL}{RTL}{Register Transfer Level}
\newacronym{SAT}{SAT}{Satisfiability}
\newacronym{SECDED}{SECDED}{Single Error Correction and Double Error Detection}
\newacronym{SEooC}{SEooC}{Safety Element out of Context}
\newacronym{SoC}{SoC}{System-on-Chip}
\newacronym{SST}{SST}{State-Space Tunneling}
\newacronym{SV}{SV}{SystemVerilog}
\newacronym{SVA}{SVA}{SystemVerilog Assertions}
\newacronym{UVM}{UVM}{Universal Verification Methodology}
\newacronym{UVC}{UVC}{UVM Verification Component}
\newacronym{VHSIC}{VHSIC}{Very High Speed Integrated Circuit}
\newacronym{VHDL}{VHDL}{VHSIC Hardware Description Language}
\newacronym{vPlan}{vPlan}{Verification Plan}
\newacronym{VSIF}{VSIF}{Verification Session Input File}
\newacronym{XML}{XML}{Extensible Markup Language}
\tikzstyle{startstop} = [rectangle, rounded corners, minimum width=3cm, minimum height=1cm,text centered, draw=black, fill=red!30]
\tikzstyle{io} = [trapezium, trapezium left angle=70, trapezium right angle=110, minimum width=3cm, minimum height=1cm, text centered, draw=black, fill=blue!30]
\tikzstyle{process} = [rectangle, minimum width=3cm, minimum height=1cm, text centered, text width=4cm, draw=black, fill=orange!30]
\tikzstyle{decision} = [diamond, minimum width=3cm, minimum height=1cm, text centered, draw=black, fill=green!30]
\tikzstyle{arrow} = [thick,->,>=stealth]

\begin{document}

\lstdefinelanguage{Verilog}{morekeywords={accept_on,alias,always,always_comb,always_ff,always_latch,and,assert,assign,assume,automatic,before,begin,bind,bins,binsof,bit,break,buf,bufif0,bufif1,byte,case,casex,casez,cell,chandle,checker,class,clocking,cmos,config,const,constraint,context,continue,cover,covergroup,coverpoint,cross,deassign,default,defparam,design,disable,dist,do,edge,else,end,endcase,endchecker,endclass,endclocking,endconfig,endfunction,endgenerate,endgroup,endinterface,endmodule,endpackage,endprimitive,endprogram,endproperty,endspecify,endsequence,endtable,endtask,enum,event,eventually,expect,export,extends,extern,final,first_match,for,force,foreach,forever,fork,forkjoin,function,generate,genvar,global,highz0,highz1,if,iff,ifnone,ignore_bins,illegal_bins,implements,implies,import,incdir,include,initial,inout,input,inside,instance,int,integer,interconnect,interface,intersect,join,join_any,join_none,large,let,liblist,library,local,localparam,logic,longint,macromodule,matches,medium,modport,module,nand,negedge,nettype,new,nexttime,nmos,nor,noshowcancelled,not,notif0,notif1,null,or,output,package,packed,parameter,pmos,posedge,primitive,priority,program,property,protected,pull0,pull1,pulldown,pullup,pulsestyle_ondetect,pulsestyle_onevent,pure,rand,randc,randcase,randsequence,rcmos,real,realtime,ref,reg,reject_on,release,repeat,restrict,return,rnmos,rpmos,rtran,rtranif0,rtranif1,s_always,s_eventually,s_nexttime,s_until,s_until_with,scalared,sequence,shortint,shortreal,showcancelled,signed,small,soft,solve,specify,specparam,static,string,strong,strong0,strong1,struct,super,supply0,supply1,sync_accept_on,sync_reject_on,table,tagged,task,this,throughout,time,timeprecision,timeunit,tran,tranif0,tranif1,tri,tri0,tri1,triand,trior,trireg,type,typedef,union,unique,unique0,unsigned,until,until_with,untyped,use,uwire,var,vectored,virtual,void,wait,wait_order,wand,weak,weak0,weak1,while,wildcard,wire,with,within,wor,xnor,xor,`uvm_create, `uvm_rand_send_with},morecomment=[l]{//}}

\title{Pragmatic Formal Verification of Sequential Error Detection and Correction Codes (ECCs) used in Safety-Critical Design}

\author{\IEEEauthorblockN{Aman Kumar}
\IEEEauthorblockA{Infineon Technologies \\
Dresden, Germany \\
Aman.Kumar@infineon.com}
}

\maketitle

\begin{abstract}
\textbf{Error Detection and Correction Codes (ECCs) are often used in digital designs to protect data integrity. Especially in safety-critical systems such as automotive electronics, ECCs are widely used and the verification of such complex logic becomes more critical considering the ISO 26262 safety standards. Exhaustive verification of ECC using formal methods has been a challenge given the high number of data bits to protect. As an example, for an ECC of 128 data bits with a possibility to detect up to four-bit errors, the combination of bit errors is given by $\binom{128}{4}+\binom{128}{3}+\binom{128}{2}+\binom{128}{1}\approx 1.1$x$10^7$. This vast analysis space often leads to bounded proof results. Moreover, the complexity and state-space increase further if the ECC has sequential encoding and decoding stages. To overcome such problems and sign-off the design with confidence within reasonable proof time, we present a pragmatic formal verification approach of complex ECC cores with several complexity reduction techniques and know-how that were learnt during the course of verification. We discuss using the linearity of the syndrome generator as a helper assertion, using the abstract model as glue logic to compare the RTL with the sequential version of the circuit, k-induction-based model checking and using mathematical relations captured as properties to simplify the verification in order to get an unbounded proof result within 24 hours of proof runtime.}
\end{abstract}


\section{Introduction}
\begin{figure}[h!]
\centering
\begin{tikzpicture}
\begin{axis}[
    xlabel={Percentage of ASIC project time spent in verification},
    ylabel={Percentage of design projects},
    xmin=10, xmax=90,
    ymin=0, ymax=30,
    xtick={10,20,30,40,50,60,70,80,90},
    ytick={0,5,10,15,20,25,30},
    legend pos=north west,
    ymajorgrids=true,
    grid style=dashed,
]

\addplot[
    color=cyan,
    mark=triangle,
    line width=0.25mm,
    ]
    coordinates {
    (20,4)(30,9)(40,11)(50,15)(60,17.5)(70,24)(80,12.5)(90,7)
    };
    \addlegendentry{2010}

\addplot[
    color=darkgray,
    mark=square,
    line width=0.25mm,
    ]
    coordinates {
    (20,3.5)(30,7.2)(40,10.5)(50,14.55)(60,20)(70,24)(80,13)(90,9.25)
    };
    \addlegendentry{2014}

\addplot[
    color=ForestGreen,
    mark=*,
    line width=0.25mm,
    ]
    coordinates {
    (20,5.8)(30,7.7)(40,12.5)(50,17)(60,19.5)(70,22)(80,10.5)(90,5)
    };
    \addlegendentry{2018}

\addplot[
    color=BurntOrange,
    mark=+,
    line width=0.25mm,
    ]
    coordinates {
    (20,6.5)(30,7)(40,14)(50,18)(60,17.5)(70,21)(80,8.5)(90,7.8)
    };
    \addlegendentry{2022}

\end{axis}
\end{tikzpicture}
\caption{Verification efforts required in overall product development \cite{VerStudy}}
\label{asic_ver_time}
\end{figure}
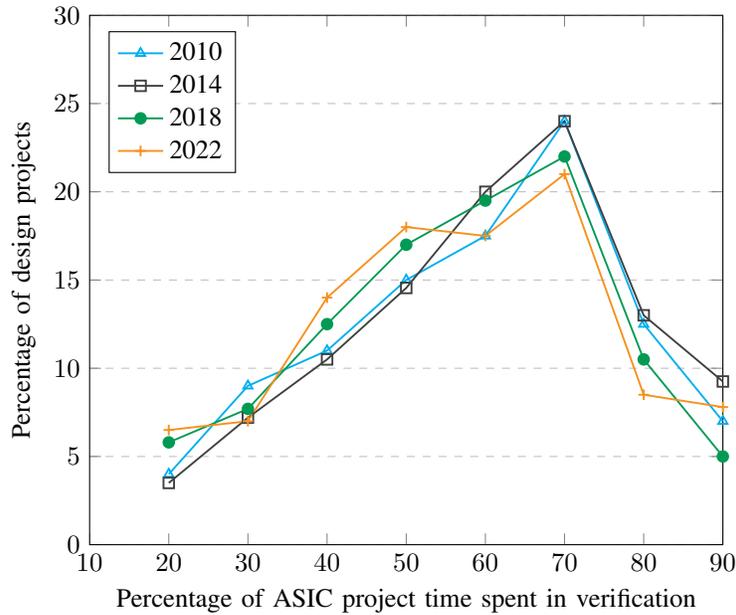

Modern \acrshort{SoC} designs are becoming more and more complex due to factors such as technology scaling, mixed-signal designs, safety and security-critical devices, more demand by customers from a single chip and many more. When such complex designs are made, one of the important aspect is to ensure data integrity. Especially in safety-critical systems and automotive electronics, the ISO 26262 functional safety standard plays a vital role to ensure the data loss is inevident. In a recent study done by the Wilson Research Group, \SI{44}{\percent} of the \acrfull{ASIC} projects are working on safety critical designs \cite{VerStudy}. Another study from the same group in Fig.~\ref{asic_ver_time} shows that verification consumes approximately a median of \SI{60}{\percent} of the overall project time. It can be depicted from the figure that the percentage of projects spending more than \SI{50}{\percent} of their time in verification increased in 2022 compared to 2018. It becomes more evident that efficient methods of verification can be used in order to make sure that such safety critical designs are exhaustively verified. Formal verification is one such method that uses mathematical proofs to verify designs in a brute-force approach.

To protect the data against soft errors such as radiation errors, electrical glitches or electro-magnetic interference \cite{soft_error1}\cite{soft_error2}, there could be many possibilities of using safety measures such as dual/triple modular redundancy (DMR/TMR), \acrshort{ECC} and alarm systems. However, the \acrshort{ECC} serves as the better option over other measures since the efficiency is higher due to less redundant flops and consequently less area required for the chip. The \acrshort{ECC} is also more flexible in terms of ability to distinguish between correctable and uncorrectable errors over other safety measures. In the context of functional safety, ISO 26262-5 also prescribes to use block replication or an \acrshort{ECC} specifically for memories \cite{iso}. On the other hand, the complexity of \acrshort{ECC} designs is rather high since there are lots of analysis-space and XOR gates. Conventionally, they are not formal friendly due to large vectors. Our paper is focused on overcoming the challenges of formal verification of such complex \acrshort{ECC} designs.

\section{Background}

\subsection{\acrfull{ECC}}

\tikzset{every picture/.style={line width=1pt}} 
\begin{figure}[h!]
\centering
\begin{tikzpicture}[x=0.75pt,y=0.75pt,yscale=-1,xscale=1]

\draw   (118,91.2) .. controls (118,80.04) and (127.04,71) .. (138.2,71) -- (198.8,71) .. controls (209.96,71) and (219,80.04) .. (219,91.2) -- (219,202.8) .. controls (219,213.96) and (209.96,223) .. (198.8,223) -- (138.2,223) .. controls (127.04,223) and (118,213.96) .. (118,202.8) -- cycle ;
\draw   (449,91.2) .. controls (449,80.04) and (458.04,71) .. (469.2,71) -- (529.8,71) .. controls (540.96,71) and (550,80.04) .. (550,91.2) -- (550,202.8) .. controls (550,213.96) and (540.96,223) .. (529.8,223) -- (469.2,223) .. controls (458.04,223) and (449,213.96) .. (449,202.8) -- cycle ;
\draw    (239,141.5) -- (276.5,141.5)(239,144.5) -- (276.5,144.5) ;
\draw [shift={(285.5,143)}, rotate = 180] [fill={rgb, 255:red, 0; green, 0; blue, 0 }  ][line width=0.08]  [draw opacity=0] (8.93,-4.29) -- (0,0) -- (8.93,4.29) -- cycle    ;
\draw    (240.5,113) -- (240.5,173)(237.5,113) -- (237.5,173) ;
\draw    (219,111.5) -- (239,111.5)(219,114.5) -- (239,114.5) ;
\draw    (219,171.5) -- (239,171.5)(219,174.5) -- (239,174.5) ;
\draw    (429,111.5) -- (440,111.5)(429,114.5) -- (440,114.5) ;
\draw [shift={(449,113)}, rotate = 180] [fill={rgb, 255:red, 0; green, 0; blue, 0 }  ][line width=0.08]  [draw opacity=0] (8.93,-4.29) -- (0,0) -- (8.93,4.29) -- cycle    ;
\draw    (429,171.5) -- (440,171.5)(429,174.5) -- (440,174.5) ;
\draw [shift={(449,173)}, rotate = 180] [fill={rgb, 255:red, 0; green, 0; blue, 0 }  ][line width=0.08]  [draw opacity=0] (8.93,-4.29) -- (0,0) -- (8.93,4.29) -- cycle    ;
\draw    (430.5,113) -- (430.5,173)(427.5,113) -- (427.5,173) ;
\draw    (386,141.5) -- (429,141.5)(386,144.5) -- (429,144.5) ;
\draw    (79,145.5) -- (109,145.5)(79,148.5) -- (109,148.5) ;
\draw [shift={(118,147)}, rotate = 180] [fill={rgb, 255:red, 0; green, 0; blue, 0 }  ][line width=0.08]  [draw opacity=0] (8.93,-4.29) -- (0,0) -- (8.93,4.29) -- cycle    ;
\draw    (550,85.2) -- (586,85.2) ;
\draw [shift={(589,85.2)}, rotate = 180] [fill={rgb, 255:red, 0; green, 0; blue, 0 }  ][line width=0.08]  [draw opacity=0] (8.93,-4.29) -- (0,0) -- (8.93,4.29) -- cycle    ;
\draw    (550,104.2) -- (586,104.2) ;
\draw [shift={(589,104.2)}, rotate = 180] [fill={rgb, 255:red, 0; green, 0; blue, 0 }  ][line width=0.08]  [draw opacity=0] (8.93,-4.29) -- (0,0) -- (8.93,4.29) -- cycle    ;
\draw    (550,122.2) -- (586,122.2) ;
\draw [shift={(589,122.2)}, rotate = 180] [fill={rgb, 255:red, 0; green, 0; blue, 0 }  ][line width=0.08]  [draw opacity=0] (8.93,-4.29) -- (0,0) -- (8.93,4.29) -- cycle    ;
\draw    (550,140.2) -- (586,140.2) ;
\draw [shift={(589,140.2)}, rotate = 180] [fill={rgb, 255:red, 0; green, 0; blue, 0 }  ][line width=0.08]  [draw opacity=0] (8.93,-4.29) -- (0,0) -- (8.93,4.29) -- cycle    ;
\draw    (550,160.2) -- (586,160.2) ;
\draw [shift={(589,160.2)}, rotate = 180] [fill={rgb, 255:red, 0; green, 0; blue, 0 }  ][line width=0.08]  [draw opacity=0] (8.93,-4.29) -- (0,0) -- (8.93,4.29) -- cycle    ;
\draw    (550,177.7) -- (580,177.7)(550,180.7) -- (580,180.7) ;
\draw [shift={(589,179.2)}, rotate = 180] [fill={rgb, 255:red, 0; green, 0; blue, 0 }  ][line width=0.08]  [draw opacity=0] (8.93,-4.29) -- (0,0) -- (8.93,4.29) -- cycle    ;
\draw    (550,201.3) -- (580,201.3)(550,204.3) -- (580,204.3) ;
\draw [shift={(589,202.8)}, rotate = 180] [fill={rgb, 255:red, 0; green, 0; blue, 0 }  ][line width=0.08]  [draw opacity=0] (8.93,-4.29) -- (0,0) -- (8.93,4.29) -- cycle    ;
\draw    (261,67.2) -- (261,118.8) ;
\draw [shift={(261,120.8)}, rotate = 270] [color={rgb, 255:red, 0; green, 0; blue, 0 }  ][line width=0.75]    (10.93,-3.29) .. controls (6.95,-1.4) and (3.31,-0.3) .. (0,0) .. controls (3.31,0.3) and (6.95,1.4) .. (10.93,3.29)   ;
\draw   (285,91.2) .. controls (285,80.04) and (294.04,71) .. (305.2,71) -- (365.8,71) .. controls (376.96,71) and (386,80.04) .. (386,91.2) -- (386,202.8) .. controls (386,213.96) and (376.96,223) .. (365.8,223) -- (305.2,223) .. controls (294.04,223) and (285,213.96) .. (285,202.8) -- cycle ;
\draw    (285,91.2) -- (386,91.2) ;
\draw    (285,109.2) -- (386,109.2) ;

\draw (154,125) node [anchor=north west][inner sep=0.75pt]   [align=left] {ECC};
\draw (144,148) node [anchor=north west][inner sep=0.75pt]   [align=left] {Encoder};
\draw (484,125) node [anchor=north west][inner sep=0.75pt]   [align=left] {ECC};
\draw (474,148) node [anchor=north west][inner sep=0.75pt]   [align=left] {Decoder};
\draw (35,133) node [anchor=north west][inner sep=0.75pt]  [font=\footnotesize] [align=left] {data\_i\\$[127:0]$};
\draw (595,77) node [anchor=north west][inner sep=0.75pt]  [font=\footnotesize] [align=left] {no\_err};
\draw (598,97) node [anchor=north west][inner sep=0.75pt]  [font=\footnotesize] [align=left] {err\_1};
\draw (599,115) node [anchor=north west][inner sep=0.75pt]  [font=\footnotesize] [align=left] {err\_2};
\draw (599,135) node [anchor=north west][inner sep=0.75pt]  [font=\footnotesize] [align=left] {err\_3};
\draw (599,152) node [anchor=north west][inner sep=0.75pt]  [font=\footnotesize] [align=left] {err\_4};
\draw (594,166) node [anchor=north west][inner sep=0.75pt]  [font=\footnotesize] [align=left] {data\_o\\$[127:0]$};
\draw (597,198) node [anchor=north west][inner sep=0.75pt]  [font=\footnotesize] [align=left] {ecc\_o\\$[24:0]$};
\draw (205,51.2) node [anchor=north west][inner sep=0.75pt]  [font=\footnotesize] [align=left] {codeword (data+ecc)};
\draw (247,123) node [anchor=north west][inner sep=0.75pt]  [font=\footnotesize] [align=left] {cw\_o};
\draw (393,123) node [anchor=north west][inner sep=0.75pt]  [font=\footnotesize] [align=left] {cw\_i};
\draw (401,75) node [anchor=north west][inner sep=0.75pt]  [font=\footnotesize] [align=left] {data\_i\\$[127:0]$};
\draw (403,182) node [anchor=north west][inner sep=0.75pt]  [font=\footnotesize] [align=left] {ecc\_i\\$[24:0]$};
\draw (309,140) node [anchor=north west][inner sep=0.75pt]   [align=left] {Memory};
\draw (289,94) node [anchor=north west][inner sep=0.75pt]   [align=left] {1010101010111};

\end{tikzpicture}
\caption{\acrshort{ECC} design with a memory}
\label{ecc}
\end{figure}
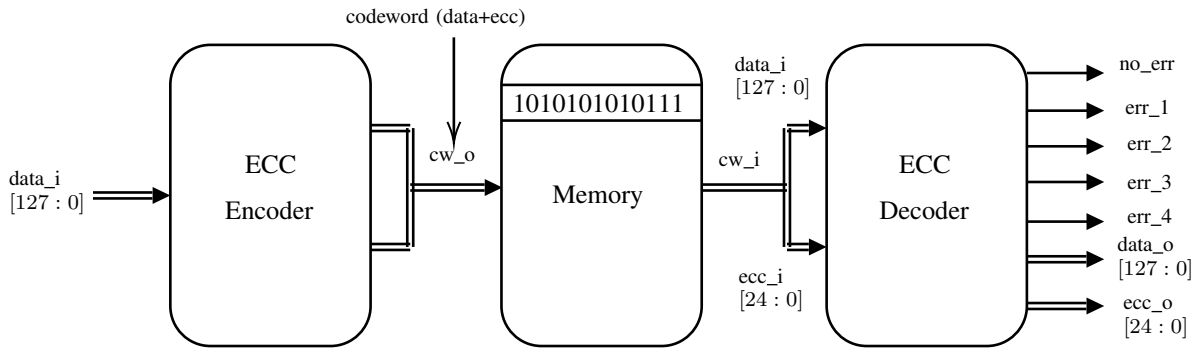

\acrfull{ECC} are used in digital designs to protect the data integrity. An ECC circuit consists of two stages, the first one is the \acrshort{ECC} encoder and second one is the \acrshort{ECC} decoder as shown in Fig.~\ref{ecc}. The \acrshort{ECC} encoder takes data input and computes the additional redundant bits called as \acrshort{ECC} bits. The \acrshort{ECC} bits are computed on two bases: the number of error bits to be detected and corrected, as well as the type of algorithm being used. Some of the algorithms such as Hamming Codes, Hsiao Codes, Reed-Solomon Codes and Bose-Chaudhuri-Hocquenghem Codes are commonly used encoding schemes for memory elements \cite{ecc_book} \cite{ecc_paper}. The resulting output from the \acrshort{ECC} encoder is a valid codeword that is written into the memory. Some soft errors can occur in the memory due to factors such as radiation errors, electrical glitches or magnetic interference that could corrupt the codeword by a bit flip resulting in an invalid codeword. The output from the memory goes to the input of the \acrshort{ECC} decoder as an invalid codeword and the \acrshort{ECC} decoder detects the bit flips as well as correct them to send out valid/corrected data. The \acrshort{ECC} used in this work is a combinatorial \acrfull{QEDTEC} \acrshort{ECC} and later, a sequentially pipelined encoding/decoding stages is also used.

The existing approach uses simulation based verification with Specman-e testbench. The verification effort takes around 6 weeks to verify the \acrshort{ECC} design. Even after several regression runs, the simulation based approach did not verify the design for all legal inputs which leads to a non brute-force approach of verification. As a result, the confidence on the design is low and not preferred for safety-critical designs. To overcome such issues, a formal based verification approach is used to exhaustively verify the correctness of the design.

\subsection{Formal Verification}
Formal verification uses technologies that mathematically analyze the space of possible behaviours of a design, rather than computing results for particular values \cite{fvbook}. It is an exhaustive verification technique that uses mathematical proof methods to verify whether the design implementation matches design specifications.

\tikzset{every picture/.style={line width=1pt}} 
\begin{figure}[h!]
\centering
\begin{tikzpicture}[x=0.75pt,y=0.75pt,yscale=-1,xscale=1]

\draw [fill={rgb, 255:red, 155; green, 155; blue, 155 }  ,fill opacity=0.1 ]  (271,133.9) .. controls (271,120.7) and (281.7,110) .. (294.9,110) -- (366.6,110) .. controls (379.8,110) and (390.5,120.7) .. (390.5,133.9) -- (390.5,266.1) .. controls (390.5,279.3) and (379.8,290) .. (366.6,290) -- (294.9,290) .. controls (281.7,290) and (271,279.3) .. (271,266.1) -- cycle ;
\draw [fill={rgb, 255:red, 255; green, 255; blue, 255 }  ,fill opacity=1 ] (281,134) .. controls (281,126.27) and (287.27,120) .. (295,120) -- (366.5,120) .. controls (374.23,120) and (380.5,126.27) .. (380.5,134) -- (380.5,176) .. controls (380.5,183.73) and (374.23,190) .. (366.5,190) -- (295,190) .. controls (287.27,190) and (281,183.73) .. (281,176) -- cycle ;
\draw [fill={rgb, 255:red, 255; green, 255; blue, 255 }  ,fill opacity=1 ] (281,225) .. controls (281,217.27) and (287.27,211) .. (295,211) -- (366.5,211) .. controls (374.23,211) and (380.5,217.27) .. (380.5,225) -- (380.5,267) .. controls (380.5,274.73) and (374.23,281) .. (366.5,281) -- (295,281) .. controls (287.27,281) and (281,274.73) .. (281,267) -- cycle ;
\draw   (156,140) .. controls (156,134.48) and (160.48,130) .. (166,130) -- (230.5,130) .. controls (236.02,130) and (240.5,134.48) .. (240.5,140) -- (240.5,170) .. controls (240.5,175.52) and (236.02,180) .. (230.5,180) -- (166,180) .. controls (160.48,180) and (156,175.52) .. (156,170) -- cycle ;
\draw   (156,230) .. controls (156,224.48) and (160.48,220) .. (166,220) -- (230.5,220) .. controls (236.02,220) and (240.5,224.48) .. (240.5,230) -- (240.5,260) .. controls (240.5,265.52) and (236.02,270) .. (230.5,270) -- (166,270) .. controls (160.48,270) and (156,265.52) .. (156,260) -- cycle ;
\draw    (240.5,155) -- (268.5,155) ;
\draw [shift={(270.5,155)}, rotate = 180] [color={rgb, 255:red, 0; green, 0; blue, 0 }  ][line width=0.75]    (10.93,-3.29) .. controls (6.95,-1.4) and (3.31,-0.3) .. (0,0) .. controls (3.31,0.3) and (6.95,1.4) .. (10.93,3.29)   ;
\draw    (240.5,244) -- (268.5,244) ;
\draw [shift={(270.5,244)}, rotate = 180] [color={rgb, 255:red, 0; green, 0; blue, 0 }  ][line width=0.75]    (10.93,-3.29) .. controls (6.95,-1.4) and (3.31,-0.3) .. (0,0) .. controls (3.31,0.3) and (6.95,1.4) .. (10.93,3.29)   ;
\draw   (176.89,135) -- (183.69,135) .. controls (187.43,135) and (190.48,138.79) .. (190.48,143.45) .. controls (190.48,148.11) and (187.43,151.9) .. (183.69,151.9) -- (176.89,151.9) -- (176.89,135) -- cycle (172.36,137.82) -- (176.89,137.82) (172.36,149.08) -- (176.89,149.08) (193.2,143.45) -- (196.82,143.45) (190.48,143.45) .. controls (190.48,142.52) and (191.09,141.76) .. (191.84,141.76) .. controls (192.59,141.76) and (193.2,142.52) .. (193.2,143.45) .. controls (193.2,144.38) and (192.59,145.14) .. (191.84,145.14) .. controls (191.09,145.14) and (190.48,144.38) .. (190.48,143.45) -- cycle ;
\draw   (204.44,147.11) -- (210.1,147.11) .. controls (214.05,147.26) and (217.58,150.56) .. (219.16,155.56) .. controls (217.58,160.57) and (214.05,163.86) .. (210.1,164.01) -- (204.44,164.01) .. controls (206.87,158.78) and (206.87,152.34) .. (204.44,147.11) -- cycle (201.04,149.93) -- (205.57,149.93) (201.04,161.2) -- (205.57,161.2) (221.88,155.56) -- (225.5,155.56) (219.16,155.56) .. controls (219.16,154.63) and (219.77,153.87) .. (220.52,153.87) .. controls (221.27,153.87) and (221.88,154.63) .. (221.88,155.56) .. controls (221.88,156.5) and (221.27,157.25) .. (220.52,157.25) .. controls (219.77,157.25) and (219.16,156.5) .. (219.16,155.56) -- cycle ;
\draw   (176.89,158.1) -- (184.23,158.1) .. controls (188.28,158.1) and (191.57,161.89) .. (191.57,166.55) .. controls (191.57,171.21) and (188.28,175) .. (184.23,175) -- (176.89,175) -- (176.89,158.1) -- cycle (172,160.92) -- (176.89,160.92) (172,172.18) -- (176.89,172.18) (191.57,166.55) -- (196.46,166.55) ;
\draw    (201.04,161.2) -- (196.46,166.55) ;
\draw    (201.04,149.93) -- (196.82,143.45) ;

\draw    (390.5,155) -- (428.5,155) ;
\draw [shift={(430.5,155)}, rotate = 180] [color={rgb, 255:red, 0; green, 0; blue, 0 }  ][line width=0.75]    (10.93,-3.29) .. controls (6.95,-1.4) and (3.31,-0.3) .. (0,0) .. controls (3.31,0.3) and (6.95,1.4) .. (10.93,3.29)   ;
\draw    (390.5,244) -- (428.5,244) ;
\draw [shift={(430.5,244)}, rotate = 180] [color={rgb, 255:red, 0; green, 0; blue, 0 }  ][line width=0.75]    (10.93,-3.29) .. controls (6.95,-1.4) and (3.31,-0.3) .. (0,0) .. controls (3.31,0.3) and (6.95,1.4) .. (10.93,3.29)   ;
\draw    (475.5,155) -- (513.5,155) ;
\draw [shift={(515.5,155)}, rotate = 180] [color={rgb, 255:red, 0; green, 0; blue, 0 }  ][line width=0.75]    (10.93,-3.29) .. controls (6.95,-1.4) and (3.31,-0.3) .. (0,0) .. controls (3.31,0.3) and (6.95,1.4) .. (10.93,3.29)   ;
\draw    (475.5,244) -- (513.5,244) ;
\draw [shift={(515.5,244)}, rotate = 180] [color={rgb, 255:red, 0; green, 0; blue, 0 }  ][line width=0.75]    (10.93,-3.29) .. controls (6.95,-1.4) and (3.31,-0.3) .. (0,0) .. controls (3.31,0.3) and (6.95,1.4) .. (10.93,3.29)   ;
\draw (585,252) node  {\includegraphics[width=83.63pt,height=52.5pt]{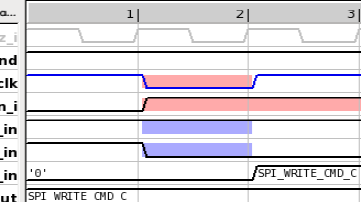}};

\draw (292,194) node [anchor=north west][inner sep=0.75pt]  [font=\small] [align=left] {Formal Verifier};
\draw (292,137) node [anchor=north west][inner sep=0.75pt]   [align=left] {Mathematical};
\draw (287,159) node [anchor=north west][inner sep=0.75pt]   [align=left] {Model of DUV};
\draw (294,229) node [anchor=north west][inner sep=0.75pt]   [align=left] {Properties as};
\draw (304,250) node [anchor=north west][inner sep=0.75pt]   [align=left] {Formulas};
\draw (183,114) node [anchor=north west][inner sep=0.75pt]  [font=\footnotesize] [align=left] {DUV};
\draw (161,224) node [anchor=north west][inner sep=0.75pt]  [font=\scriptsize] [align=left] {SVA\_AST:\\assert property (\\@(posedge clk)\\error $|$=$>$ reg\_bit);};
\draw (173,204) node [anchor=north west][inner sep=0.75pt]  [font=\footnotesize] [align=left] {Properties};
\draw (433,148) node [anchor=north west][inner sep=0.75pt]   [align=left] {\textcolor{ForestGreen}{PASS}};
\draw (437,237) node [anchor=north west][inner sep=0.75pt]   [align=left] {\textcolor{red}{FAIL}};
\draw (518,148) node [anchor=north west][inner sep=0.75pt]   [align=left] {\textcolor{ForestGreen}{PROVEN}};
\draw (518,200) node [anchor=north west][inner sep=0.75pt]   [align=left] {\textcolor{red}{COUNTER EXAMPLE}};

\end{tikzpicture}
\caption{Formal verifier \cite{aman_dvcon_21}}
\label{formal_verifier}
\end{figure}
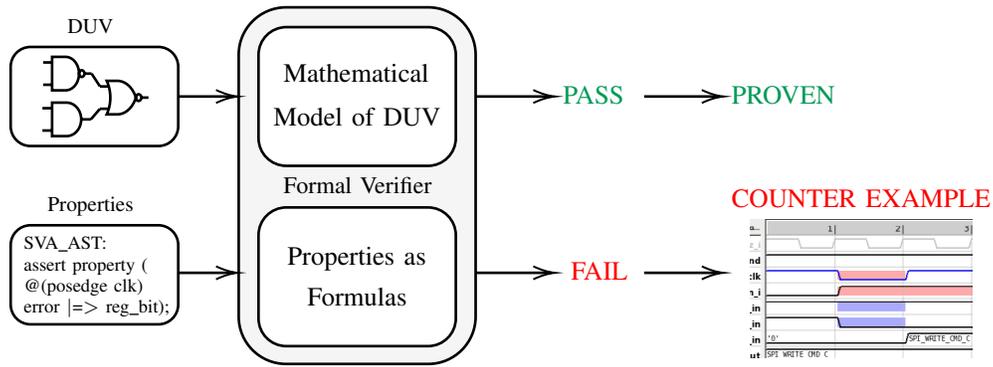

Fig.~\ref{formal_verifier} shows the working of a formal verifier. There are two inputs to the formal verifier tool. On the one hand, the \acrfull{DUV} is fed into the tool and converted into a mathematical model. On the other hand, properties, written in \acrfull{SVA} that capture the intent of the design are also fed into the tool. The tool then converts these properties into mathematical formulas. In the next step, the tool tries to prove these mathematical formulas on the mathematical model of the \acrshort{DUV}. If the properties do not hold, it is said to be failed, and a \acrfull{CEX} is generated by the tool to debug further. In general, the absence of a \acrshort{CEX} is a pass or proven result. Formal verification uses clever algorithms to verify the functional correctness of the design exhaustively \cite{fvbook}. However, for larger designs, it suffers state-space explosion and often takes longer than reasonable evaluation times for a conclusive proof result \cite{fvbook}.

\section{Formal Verification of \acrshort{ECC}}
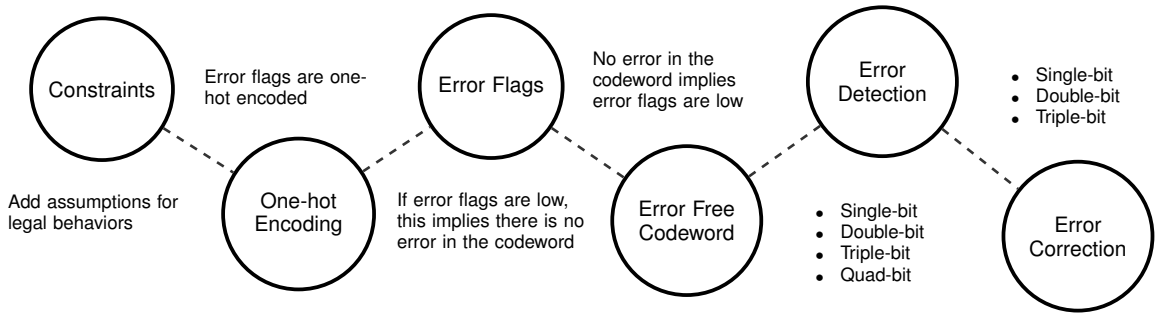
\begin{figure}[h!]
\centering
\begin{tikzpicture}[
 bigcircle/.style={ 
    text width=1.6cm, 
    align=center, 
    line width=0.5mm, 
    draw, 
    circle, 
    font=\sffamily\footnotesize 
  },
 desc/.style 2 args={ 
  text width=2.5cm, 
  font=\sffamily\scriptsize\RaggedRight, 
  label={[#1,yshift=-1.5ex,font=\sffamily\footnotesize]above:#2} 
  },
 node distance=3mm and 2mm 
]

\node [bigcircle] (circ1) {Constraints};
\node [desc={black}{},below=of circ1] (list1) {
Add assumptions for legal behaviors
};

\node [bigcircle,black,right=of list1] (circ2) {One-hot Encoding};
\node [desc={black}{},above=of circ2] (list2) {
Error flags are one-hot encoded
};

\node [bigcircle,black,right=of list2] (circ3) {Error Flags};
\node [desc={black}{},below=of circ3] (list3) {
If error flags are low, this implies there is no error in the codeword
};

\node [bigcircle,black,right=of list3] (circ4) {Error Free Codeword};
\node [desc={black}{},above=of circ4] (list4) {
No error in the codeword implies error flags are low
};

\node [bigcircle,black,right=of list4] (circ5) {Error Detection};
\node [desc={black}{},below=of circ5] (list5) {
\begin{itemize}
\setlength\itemsep{0pt}
\item Single-bit
\item Double-bit
\item Triple-bit
\item Quad-bit
\end{itemize}
};

\node [bigcircle,black,right=of list5] (circ6) {Error Correction};
\node [desc={black}{},above=of circ6] (h) {
\begin{itemize}
\setlength\itemsep{0pt}
\item Single-bit
\item Double-bit
\item Triple-bit
\end{itemize}
};

\draw [dashed,black!80] (circ1) -- (circ2) -- (circ3) -- (circ4) -- (circ5) -- (circ6);
\end{tikzpicture}
\caption{Verification plan for ECC}
\label{vplan}
\end{figure}

The motivation to use a formal based verification approach is to exhaustively verify that the design implementation matches the design specifications. However, since the \acrshort{ECC} used in this work is a large design with a vast analysis space to cover, several complexity reduction techniques are used as well. To understand the formal verification, a step-wise approach is used and explained. At first, a verification plan is prepared as mentioned in the Fig.~\ref{vplan}. Based on the verification plan, different properties are written to capture the intent of the design and prove the correctness.

\subsection{\textbf{Addressing Complexity Step 1: Remove Memory and Model Errors}}
The memory is irrelevant for proving the correctness of ECCs. In fact, memories are not very formal friendly and would increase the state-space, increase complexity as well as increase the proof time for the formal tool \cite{formal_friendly}. As a result, memory can be safely removed from the analysis space.

\tikzset{every picture/.style={line width=1pt}} 
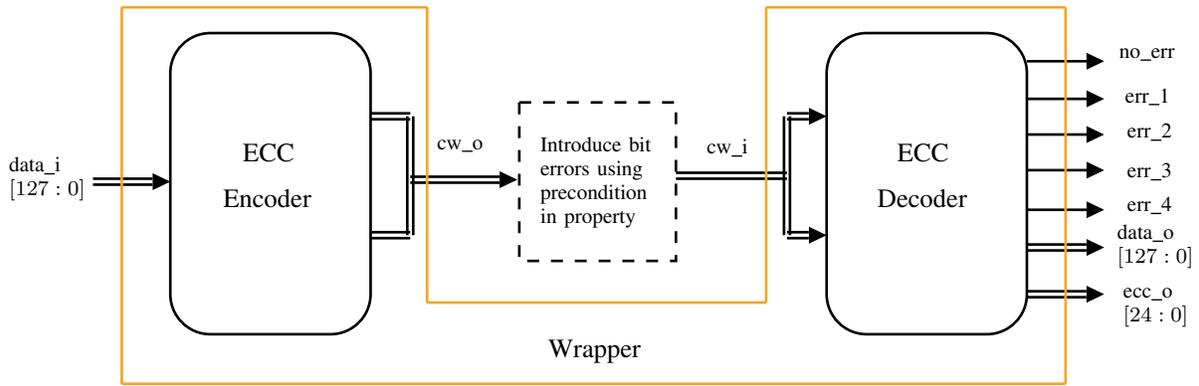
\begin{figure}[h!]
\centering
\begin{tikzpicture}[x=0.75pt,y=0.75pt,yscale=-1,xscale=1]

\draw   (125,84.2) .. controls (125,73.04) and (134.04,64) .. (145.2,64) -- (205.8,64) .. controls (216.96,64) and (226,73.04) .. (226,84.2) -- (226,195.8) .. controls (226,206.96) and (216.96,216) .. (205.8,216) -- (145.2,216) .. controls (134.04,216) and (125,206.96) .. (125,195.8) -- cycle ;
\draw   (456,84.2) .. controls (456,73.04) and (465.04,64) .. (476.2,64) -- (536.8,64) .. controls (547.96,64) and (557,73.04) .. (557,84.2) -- (557,195.8) .. controls (557,206.96) and (547.96,216) .. (536.8,216) -- (476.2,216) .. controls (465.04,216) and (456,206.96) .. (456,195.8) -- cycle ;
\draw  [dash pattern={on 4.5pt off 4.5pt}] (301,99) -- (380,99) -- (380,179) -- (301,179) -- cycle ;
\draw    (246,136.5) -- (291,136.5)(246,139.5) -- (291,139.5) ;
\draw [shift={(300,138)}, rotate = 180] [fill={rgb, 255:red, 0; green, 0; blue, 0 }  ][line width=0.08]  [draw opacity=0] (8.93,-4.29) -- (0,0) -- (8.93,4.29) -- cycle    ;
\draw    (247.5,106) -- (247.5,166)(244.5,106) -- (244.5,166) ;
\draw    (226,104.5) -- (246,104.5)(226,107.5) -- (246,107.5) ;
\draw    (226,164.5) -- (246,164.5)(226,167.5) -- (246,167.5) ;
\draw    (436,104.5) -- (447,104.5)(436,107.5) -- (447,107.5) ;
\draw [shift={(456,106)}, rotate = 180] [fill={rgb, 255:red, 0; green, 0; blue, 0 }  ][line width=0.08]  [draw opacity=0] (8.93,-4.29) -- (0,0) -- (8.93,4.29) -- cycle    ;
\draw    (436,164.5) -- (447,164.5)(436,167.5) -- (447,167.5) ;
\draw [shift={(456,166)}, rotate = 180] [fill={rgb, 255:red, 0; green, 0; blue, 0 }  ][line width=0.08]  [draw opacity=0] (8.93,-4.29) -- (0,0) -- (8.93,4.29) -- cycle    ;
\draw    (437.5,106) -- (437.5,166)(434.5,106) -- (434.5,166) ;
\draw    (380,134.5) -- (436,134.5)(380,137.5) -- (436,137.5) ;
\draw    (86,137.5) -- (116,137.5)(86,140.5) -- (116,140.5) ;
\draw [shift={(125,139)}, rotate = 180] [fill={rgb, 255:red, 0; green, 0; blue, 0 }  ][line width=0.08]  [draw opacity=0] (8.93,-4.29) -- (0,0) -- (8.93,4.29) -- cycle    ;
\draw    (557,78.2) -- (593,78.2) ;
\draw [shift={(596,78.2)}, rotate = 180] [fill={rgb, 255:red, 0; green, 0; blue, 0 }  ][line width=0.08]  [draw opacity=0] (8.93,-4.29) -- (0,0) -- (8.93,4.29) -- cycle    ;
\draw    (557,97.2) -- (593,97.2) ;
\draw [shift={(596,97.2)}, rotate = 180] [fill={rgb, 255:red, 0; green, 0; blue, 0 }  ][line width=0.08]  [draw opacity=0] (8.93,-4.29) -- (0,0) -- (8.93,4.29) -- cycle    ;
\draw    (557,115.2) -- (593,115.2) ;
\draw [shift={(596,115.2)}, rotate = 180] [fill={rgb, 255:red, 0; green, 0; blue, 0 }  ][line width=0.08]  [draw opacity=0] (8.93,-4.29) -- (0,0) -- (8.93,4.29) -- cycle    ;
\draw    (557,133.2) -- (593,133.2) ;
\draw [shift={(596,133.2)}, rotate = 180] [fill={rgb, 255:red, 0; green, 0; blue, 0 }  ][line width=0.08]  [draw opacity=0] (8.93,-4.29) -- (0,0) -- (8.93,4.29) -- cycle    ;
\draw    (557,153.2) -- (593,153.2) ;
\draw [shift={(596,153.2)}, rotate = 180] [fill={rgb, 255:red, 0; green, 0; blue, 0 }  ][line width=0.08]  [draw opacity=0] (8.93,-4.29) -- (0,0) -- (8.93,4.29) -- cycle    ;
\draw    (557,170.7) -- (587,170.7)(557,173.7) -- (587,173.7) ;
\draw [shift={(596,172.2)}, rotate = 180] [fill={rgb, 255:red, 0; green, 0; blue, 0 }  ][line width=0.08]  [draw opacity=0] (8.93,-4.29) -- (0,0) -- (8.93,4.29) -- cycle    ;
\draw    (557,194.3) -- (587,194.3)(557,197.3) -- (587,197.3) ;
\draw [shift={(596,195.8)}, rotate = 180] [fill={rgb, 255:red, 0; green, 0; blue, 0 }  ][line width=0.08]  [draw opacity=0] (8.93,-4.29) -- (0,0) -- (8.93,4.29) -- cycle    ;
\draw [color={rgb, 255:red, 245; green, 166; blue, 35 }  ,draw opacity=1 ]   (100.5,51) -- (100.5,241) ;
\draw [color={rgb, 255:red, 245; green, 166; blue, 35 }  ,draw opacity=1 ]   (254.5,51) -- (254.5,200) ;
\draw [color={rgb, 255:red, 245; green, 166; blue, 35 }  ,draw opacity=1 ]   (576.5,51) -- (576.5,241) ;
\draw [color={rgb, 255:red, 245; green, 166; blue, 35 }  ,draw opacity=1 ]   (100.5,241) -- (576.5,241) ;
\draw [color={rgb, 255:red, 245; green, 166; blue, 35 }  ,draw opacity=1 ]   (254.5,200) -- (425.5,200) ;
\draw [color={rgb, 255:red, 245; green, 166; blue, 35 }  ,draw opacity=1 ]   (100.5,51) -- (254.5,51) ;
\draw [color={rgb, 255:red, 245; green, 166; blue, 35 }  ,draw opacity=1 ]   (425.5,51) -- (576.5,51) ;
\draw [color={rgb, 255:red, 245; green, 166; blue, 35 }  ,draw opacity=1 ]   (425.5,51) -- (425.5,200) ;

\draw (160,118) node [anchor=north west][inner sep=0.75pt]   [align=left] {ECC};
\draw (150,141) node [anchor=north west][inner sep=0.75pt]   [align=left] {Encoder};
\draw (490,118) node [anchor=north west][inner sep=0.75pt]   [align=left] {ECC};
\draw (480,141) node [anchor=north west][inner sep=0.75pt]   [align=left] {Decoder};
\draw (310,115) node [anchor=north west][inner sep=0.75pt]  [font=\footnotesize] [align=left] {Introduce bit\\errors using\\precondition\\in property};
\draw (42,125) node [anchor=north west][inner sep=0.75pt]  [font=\footnotesize] [align=left] {data\_i\\$[127:0]$};
\draw (602,70) node [anchor=north west][inner sep=0.75pt]  [font=\footnotesize] [align=left] {no\_err};
\draw (605,90) node [anchor=north west][inner sep=0.75pt]  [font=\footnotesize] [align=left] {err\_1};
\draw (606,108) node [anchor=north west][inner sep=0.75pt]  [font=\footnotesize] [align=left] {err\_2};
\draw (606,128) node [anchor=north west][inner sep=0.75pt]  [font=\footnotesize] [align=left] {err\_3};
\draw (606,145) node [anchor=north west][inner sep=0.75pt]  [font=\footnotesize] [align=left] {err\_4};
\draw (601,159) node [anchor=north west][inner sep=0.75pt]  [font=\footnotesize] [align=left] {data\_o\\$[127:0]$};
\draw (604,191) node [anchor=north west][inner sep=0.75pt]  [font=\footnotesize] [align=left] {ecc\_o\\$[24:0]$};
\draw (258,116) node [anchor=north west][inner sep=0.75pt]  [font=\footnotesize] [align=left] {cw\_o};
\draw (394,116) node [anchor=north west][inner sep=0.75pt]  [font=\footnotesize] [align=left] {cw\_i};
\draw (314,217) node [anchor=north west][inner sep=0.75pt]   [align=left] {Wrapper};

\end{tikzpicture}
\caption{Remove memory and model errors}
\label{wrapper}
\end{figure}

The next step is to model the errors in \acrshort{ECC} codeword. For this purpose a \acrfull{SV} wrapper is prepared that instantiates the \acrshort{ECC} encoder and the \acrshort{ECC} decoder as shown in Fig.~\ref{wrapper}. The encoder outputs act as the primary outputs whereas the decoder inputs act as the primary inputs for the formal tool. As a result, the formal tool is free to take any possible values, unless constrained, for these primary inputs and outputs. The bit errors are introduced in the precondition of the properties written in \acrfull{SVA}.

While preparing the properties to prove the correctness of the \acrshort{ECC}, one of the requirements was to verify that exactly one of the error flags is high i.e., the error flags are one-hot encoded. Upon writing the property as mentioned in Listing~\ref{onehot}, a bug was found where this requirement was violated. Although it was fixed in the design later, this already demonstrates how formal verification comes handy to expose simple bugs that could take more effort and time if verified using the conventional simulation based approach.

\vspace{0.25cm}
\begin{lstlisting}[language=Verilog, float=h!, caption=\acrshort{SVA} property to prove error flags are one-hot encoded, basicstyle=\ttfamily, label={onehot}]
//==========================================================================
// Prove that exactly one of the error flags is high i.e., one-hot encoding
//==========================================================================
property onehot_error_flags;
    (1
  |->
    ( ($onehot({no_err, err_1, err_2, err_3, err_4})) );
endproperty
ap_onehot_error_flags : assert property(onehot_error_flags);
\end{lstlisting}

In a standard brute-force approach, a \acrshort{SVA} property to detect all 4 bit errors (multi bit errors) could be written as mentioned in Listing~\ref{mberr}. The analysis space for the formal tool to prove the property in Listing~\ref{mberr} corresponds to $2^{128} $x$ \binom{153}{4}$ which is huge and results in an unbounded proof result.
\vspace{0.25cm}
\begin{lstlisting}[language=Verilog, float=h!, caption=\acrshort{SVA} property to detect all multi bit errors, basicstyle=\ttfamily, label={mberr}]
//=================================================
// All multi bit errors (MBERR) shall be detected
//=================================================
property MBERR_detection;
    ($countones(cw_o ^ cw_i) == 4)
  |->
    (err_4 && !err_1 && !err_2 && !err_3 && !no_err);
endproperty
ap_MBERR_detection : assert property(MBERR_detection);
\end{lstlisting}

\subsection{\textbf{Addressing Complexity Step 2: The Linearity Approach}}
To address the complexity with large state-space, an attempt to explore the internals of the \acrshort{ECC} decoder was made. The decoder primarily consists of three units: the syndrome generator, the error detection unit and the error correction unit as shown in Fig.~\ref{syndrome_gen}. The syndrome generator computes the error syndrome for the codeword and is based on the encoding scheme used to generate the check bits \cite{keerthi_ecc}. The error detection unit sets the error flags and the error correction unit corrects the data in case of an invalid codeword based on the output from the syndrome generator.

\tikzset{every picture/.style={line width=1pt}} 
\begin{figure}[h!]
\centering
\begin{tikzpicture}[x=0.75pt,y=0.75pt,yscale=-1,xscale=1]

\draw   (96,104.2) .. controls (96,93.04) and (105.04,84) .. (116.2,84) -- (176.8,84) .. controls (187.96,84) and (197,93.04) .. (197,104.2) -- (197,215.8) .. controls (197,226.96) and (187.96,236) .. (176.8,236) -- (116.2,236) .. controls (105.04,236) and (96,226.96) .. (96,215.8) -- cycle ;
\draw   (379.5,113.7) .. controls (379.5,97.3) and (392.8,84) .. (409.2,84) -- (498.3,84) .. controls (514.7,84) and (528,97.3) .. (528,113.7) -- (528,206.3) .. controls (528,222.7) and (514.7,236) .. (498.3,236) -- (409.2,236) .. controls (392.8,236) and (379.5,222.7) .. (379.5,206.3) -- cycle ;
\draw  [dash pattern={on 4.5pt off 4.5pt}] (251,119) -- (330,119) -- (330,199) -- (251,199) -- cycle ;
\draw    (217,156.5) -- (243,156.5)(217,159.5) -- (243,159.5) ;
\draw [shift={(252,158)}, rotate = 180] [fill={rgb, 255:red, 0; green, 0; blue, 0 }  ][line width=0.08]  [draw opacity=0] (8.93,-4.29) -- (0,0) -- (8.93,4.29) -- cycle    ;
\draw    (218.5,126) -- (218.5,186)(215.5,126) -- (215.5,186) ;
\draw    (197,124.5) -- (217,124.5)(197,127.5) -- (217,127.5) ;
\draw    (197,184.5) -- (217,184.5)(197,187.5) -- (217,187.5) ;
\draw    (359,124.5) -- (370,124.5)(359,127.5) -- (370,127.5) ;
\draw [shift={(379,126)}, rotate = 180] [fill={rgb, 255:red, 0; green, 0; blue, 0 }  ][line width=0.08]  [draw opacity=0] (8.93,-4.29) -- (0,0) -- (8.93,4.29) -- cycle    ;
\draw    (359,184.5) -- (370,184.5)(359,187.5) -- (370,187.5) ;
\draw [shift={(379,186)}, rotate = 180] [fill={rgb, 255:red, 0; green, 0; blue, 0 }  ][line width=0.08]  [draw opacity=0] (8.93,-4.29) -- (0,0) -- (8.93,4.29) -- cycle    ;
\draw    (361.5,126) -- (361.5,186)(358.5,126) -- (358.5,186) ;
\draw    (330,154.5) -- (360,154.5)(330,157.5) -- (360,157.5) ;
\draw    (57,157.5) -- (87,157.5)(57,160.5) -- (87,160.5) ;
\draw [shift={(96,159)}, rotate = 180] [fill={rgb, 255:red, 0; green, 0; blue, 0 }  ][line width=0.08]  [draw opacity=0] (8.93,-4.29) -- (0,0) -- (8.93,4.29) -- cycle    ;
\draw    (526,101.2) -- (564,101.2) ;
\draw [shift={(567,101.2)}, rotate = 180] [fill={rgb, 255:red, 0; green, 0; blue, 0 }  ][line width=0.08]  [draw opacity=0] (8.93,-4.29) -- (0,0) -- (8.93,4.29) -- cycle    ;
\draw    (528,117.2) -- (564,117.2) ;
\draw [shift={(567,117.2)}, rotate = 180] [fill={rgb, 255:red, 0; green, 0; blue, 0 }  ][line width=0.08]  [draw opacity=0] (8.93,-4.29) -- (0,0) -- (8.93,4.29) -- cycle    ;
\draw    (528,135.2) -- (564,135.2) ;
\draw [shift={(567,135.2)}, rotate = 180] [fill={rgb, 255:red, 0; green, 0; blue, 0 }  ][line width=0.08]  [draw opacity=0] (8.93,-4.29) -- (0,0) -- (8.93,4.29) -- cycle    ;
\draw    (528,153.2) -- (564,153.2) ;
\draw [shift={(567,153.2)}, rotate = 180] [fill={rgb, 255:red, 0; green, 0; blue, 0 }  ][line width=0.08]  [draw opacity=0] (8.93,-4.29) -- (0,0) -- (8.93,4.29) -- cycle    ;
\draw    (528,173.2) -- (564,173.2) ;
\draw [shift={(567,173.2)}, rotate = 180] [fill={rgb, 255:red, 0; green, 0; blue, 0 }  ][line width=0.08]  [draw opacity=0] (8.93,-4.29) -- (0,0) -- (8.93,4.29) -- cycle    ;
\draw    (528,190.7) -- (558,190.7)(528,193.7) -- (558,193.7) ;
\draw [shift={(567,192.2)}, rotate = 180] [fill={rgb, 255:red, 0; green, 0; blue, 0 }  ][line width=0.08]  [draw opacity=0] (8.93,-4.29) -- (0,0) -- (8.93,4.29) -- cycle    ;
\draw    (528,214.3) -- (558,214.3)(528,217.3) -- (558,217.3) ;
\draw [shift={(567,215.8)}, rotate = 180] [fill={rgb, 255:red, 0; green, 0; blue, 0 }  ][line width=0.08]  [draw opacity=0] (8.93,-4.29) -- (0,0) -- (8.93,4.29) -- cycle    ;
\draw [color={rgb, 255:red, 245; green, 166; blue, 35 }  ,draw opacity=1 ]   (71.5,71) -- (71.5,261) ;
\draw [color={rgb, 255:red, 245; green, 166; blue, 35 }  ,draw opacity=1 ]   (225.5,71) -- (225.5,220) ;
\draw [color={rgb, 255:red, 245; green, 166; blue, 35 }  ,draw opacity=1 ]   (547.5,71) -- (547.5,261) ;
\draw [color={rgb, 255:red, 245; green, 166; blue, 35 }  ,draw opacity=1 ]   (71.5,261) -- (547.5,261) ;
\draw [color={rgb, 255:red, 245; green, 166; blue, 35 }  ,draw opacity=1 ]   (225.5,220) -- (342.5,220) ;
\draw [color={rgb, 255:red, 245; green, 166; blue, 35 }  ,draw opacity=1 ]   (71.5,71) -- (225.5,71) ;
\draw [color={rgb, 255:red, 245; green, 166; blue, 35 }  ,draw opacity=1 ]   (342.5,71) -- (547.5,71) ;
\draw [color={rgb, 255:red, 245; green, 166; blue, 35 }  ,draw opacity=1 ]   (342.5,71) -- (342.5,220) ;
\draw   (386,119.9) .. controls (386,113.33) and (391.33,108) .. (397.9,108) -- (433.6,108) .. controls (440.17,108) and (445.5,113.33) .. (445.5,119.9) -- (445.5,188.1) .. controls (445.5,194.67) and (440.17,200) .. (433.6,200) -- (397.9,200) .. controls (391.33,200) and (386,194.67) .. (386,188.1) -- cycle ;
\draw   (464.5,114.6) .. controls (464.5,108.75) and (469.25,104) .. (475.1,104) -- (512.9,104) .. controls (518.75,104) and (523.5,108.75) .. (523.5,114.6) -- (523.5,146.4) .. controls (523.5,152.25) and (518.75,157) .. (512.9,157) -- (475.1,157) .. controls (469.25,157) and (464.5,152.25) .. (464.5,146.4) -- cycle ;
\draw   (464.5,171.2) .. controls (464.5,167.22) and (467.72,164) .. (471.7,164) -- (516.3,164) .. controls (520.28,164) and (523.5,167.22) .. (523.5,171.2) -- (523.5,192.8) .. controls (523.5,196.78) and (520.28,200) .. (516.3,200) -- (471.7,200) .. controls (467.72,200) and (464.5,196.78) .. (464.5,192.8) -- cycle ;
\draw    (452.5,121) -- (452.5,181)(449.5,121) -- (449.5,181) ;
\draw    (450.9,179.5) -- (456.42,179.12)(451.1,182.5) -- (456.62,182.12) ;
\draw [shift={(465.5,180)}, rotate = 176.05] [fill={rgb, 255:red, 0; green, 0; blue, 0 }  ][line width=0.08]  [draw opacity=0] (8.93,-4.29) -- (0,0) -- (8.93,4.29) -- cycle    ;
\draw    (451,119.5) -- (456,119.5)(451,122.5) -- (456,122.5) ;
\draw [shift={(465,121)}, rotate = 180] [fill={rgb, 255:red, 0; green, 0; blue, 0 }  ][line width=0.08]  [draw opacity=0] (8.93,-4.29) -- (0,0) -- (8.93,4.29) -- cycle    ;
\draw    (446,149.5) -- (451,149.5)(446,152.5) -- (451,152.5) ;

\draw (132,138) node [anchor=north west][inner sep=0.75pt]   [align=left] {ECC};
\draw (122,161) node [anchor=north west][inner sep=0.75pt]   [align=left] {Encoder};
\draw (415,213) node [anchor=north west][inner sep=0.75pt]   [align=left] {ECC Decoder};
\draw (260,134) node [anchor=north west][inner sep=0.75pt]  [font=\footnotesize] [align=left] {Introduce bit\\errors using\\precondition\\in property};
\draw (13,145) node [anchor=north west][inner sep=0.75pt]  [font=\footnotesize] [align=left] {data\_i\\$[127:0]$};
\draw (573,90) node [anchor=north west][inner sep=0.75pt]  [font=\footnotesize] [align=left] {no\_err};
\draw (576,110) node [anchor=north west][inner sep=0.75pt]  [font=\footnotesize] [align=left] {err\_1};
\draw (577,128) node [anchor=north west][inner sep=0.75pt]  [font=\footnotesize] [align=left] {err\_2};
\draw (577,148) node [anchor=north west][inner sep=0.75pt]  [font=\footnotesize] [align=left] {err\_3};
\draw (577,165) node [anchor=north west][inner sep=0.75pt]  [font=\footnotesize] [align=left] {err\_4};
\draw (572,179) node [anchor=north west][inner sep=0.75pt]  [font=\footnotesize] [align=left] {data\_o\\$[127:0]$};
\draw (575,211) node [anchor=north west][inner sep=0.75pt]  [font=\footnotesize] [align=left] {ecc\_o\\$[24:0]$};
\draw (220,136) node [anchor=north west][inner sep=0.75pt]  [font=\footnotesize] [align=left] {cw\_o};
\draw (332,136) node [anchor=north west][inner sep=0.75pt]  [font=\footnotesize] [align=left] {cw\_i};
\draw (257,237) node [anchor=north west][inner sep=0.75pt]   [align=left] {Wrapper};
\draw (391,141) node [anchor=north west][inner sep=0.75pt]  [font=\footnotesize] [align=left] {Syndrome};
\draw (391,157) node [anchor=north west][inner sep=0.75pt]  [font=\footnotesize] [align=left] {Generator};
\draw (480,117) node [anchor=north west][inner sep=0.75pt]  [font=\footnotesize] [align=left] {Error};
\draw (470,132) node [anchor=north west][inner sep=0.75pt]  [font=\footnotesize] [align=left] {Detection};
\draw (480,168) node [anchor=north west][inner sep=0.75pt]  [font=\footnotesize] [align=left] {Error};
\draw (468,183) node [anchor=north west][inner sep=0.75pt]  [font=\footnotesize] [align=left] {Correction};

\end{tikzpicture}
\caption{Exploring ECC Decoder and its linearity}
\label{syndrome_gen}
\end{figure}
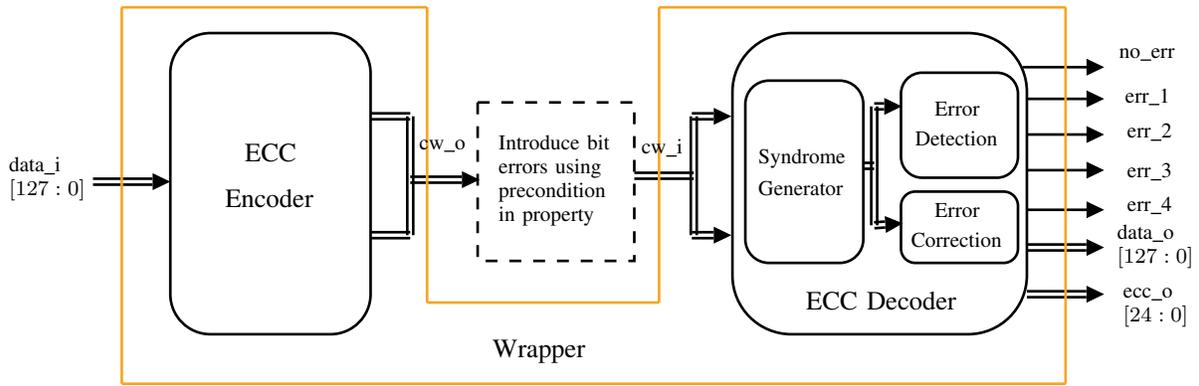

The following observations were made for the syndrome generator:
\begin{itemize}
    \item No error in codeword implies the syndrome output is a null vector.
    \item Syndrome generator is a linear function i.e., the syndrome output is independent of the data input.
    \item Syndrome output only depends on the erroneous bit positions.
\end{itemize}
A function is an algebraic linear function if it preserves the addition operation, i.e., it satisfies the following mathematical rule:
\begin{equation}
f(x) + f(y) = f(x+y)
\end{equation}
where $x, y$ are arbitrary vector spaces. A linear function preserves the vector addition and scalar multiplication irrespective of the vector spaces and the scalar value \cite{keerthi_ecc}. Furthermore, the syndrome output is observed to be the erroneous bit position when a bit flip is introduced in the valid codeword. This is confirmed by noting the syndrom output for different codeword but same bit-error position as mentioned in Table~\ref{err_bit}:

\begin{table}[h!]
\centering
\begin{tabular}{|p{3cm}|p{2.5cm}|p{2.5cm}|}
\hline
\textbf{Codeword} & \textbf{Bit-Error Position} & \textbf{Syndrome Output} \\
\hline
128'hcaba79a638aca501 & 6th bit position & 25'd6 \\
\hline
128'hec5e6af55ea3 & 6th bit position & 25'd6 \\
\hline
128'hcaba79a638aca501 & 18th bit position & 25'd18 \\
\hline
128'hec5e6af55ea3 & 18th bit position & 25'd18 \\
\hline
\end{tabular}
\caption{Syndrome output only depends on the erroneous bit positions}
\label{err_bit}
\end{table}

To prove the above three observations on the design, simple \acrshort{SVA} properties could be written as stated in Listings \ref{err_free_syn}, \ref{synd_linear} and \ref{err_bit_pos}.
\vspace{0.25cm}
\begin{lstlisting}[language=Verilog, float=h!, caption=\acrshort{SVA} property to prove syndrome output is a null vector for error free codeword, basicstyle=\ttfamily, label={err_free_syn}]
//=================================================
// Syndrome output of error free codeword is zero
//=================================================
property error_free_syndrome;
    (cw_i == cw_o) 
  |->
    (syn_o == 0);
endproperty
ap_error_free_syndrome : assert property(error_free_syndrome);
\end{lstlisting}
\vspace{0.25cm}
\begin{lstlisting}[language=Verilog, float=h!, caption=\acrshort{SVA} property to prove syndrome generator is a liner function, basicstyle=\ttfamily, label={synd_linear}]
//====================================================
// Prove that syndrome generator is a linear function
//====================================================
property syndrome_linearity;
    (syn(x) ^ syn(y) == syn (x ^ y));
endproperty
ap_syndrome_linearity : assert property(syndrome_linearity);
\end{lstlisting}
\vspace{0.25cm}
\begin{lstlisting}[language=Verilog, float=h!, caption=\acrshort{SVA} property to prove syndrome output only depends on the erroneous bit position, basicstyle=\ttfamily, label={err_bit_pos}]
//=======================================================
// Syndrome output depends on the erroneous bit position
//=======================================================
property error_bit_position;
    (($countones(cw_o ^ cw_i) == 1) &&
    (cw_o[6] != cw_i[6]))
  |->
    (syn_o == 6);
endproperty
ap_error_bit_position : assert property(error_bit_position);
\end{lstlisting}

Proving linearity means the syndrome output is independent of the data input. These helper asserts will be proven and therefore implicitly assumed for improved proof performance of other properties. Since the data input does not contribute to the syndrome output, the multi bit error detection can now be further simplified to use a randomly selected fixed data input and prove the property as shown in Listing~\ref{mberr_linear}. Similarly, a \acrshort{SVA} property to prove 3 bit error correction could be written as Listing~\ref{tberr_corr}. The reduced analysis space for the formal tool to prove the property in Listing~\ref{mberr_linear} is $1 $x$ \binom{153}{4}$ and the proof time is 10 hours.
\vspace{0.25cm}
\begin{lstlisting}[language=Verilog, float=h!, caption=\acrshort{SVA} property to to detect all multi bit errors together with linearity approach, basicstyle=\ttfamily, label={mberr_linear}]
//=================================================
// All multi bit errors (MBERR) shall be detected
//=================================================
property MBERR_detection;
    ((data_i == 128'hcaba79a638aca501) // random
    && ($countones(cw_o ^ cw_i) == 4))
  |->
    (err_4 && !err_1 && !err_2 && !err_3 && !no_err);
endproperty
ap_MBERR_detection : assert property(MBERR_detection);
\end{lstlisting}
\vspace{0.25cm}
\begin{lstlisting}[language=Verilog, float=h!, caption=\acrshort{SVA} property to to correct all 3 bit errors together with linearity approach, basicstyle=\ttfamily, label={tberr_corr}]
//==================================================
// All triple bit errors (TBERR) shall be corrected
//==================================================
property TBERR_correction;
    ((data_i == 128'hcaba79a638aca501) &&
    ($countones(cw_o ^ cw_i) == 3) && (err_3))
  |->
    ((dec_data_o == enc_data_i) && (dec_ecc_o == enc_ecc_o));
endproperty
ap_TBERR_correction : assert property(TBERR_correction);
\end{lstlisting}

\subsection{\textbf{Addressing Complexity Step 3: Using Reduced Latency Model}}
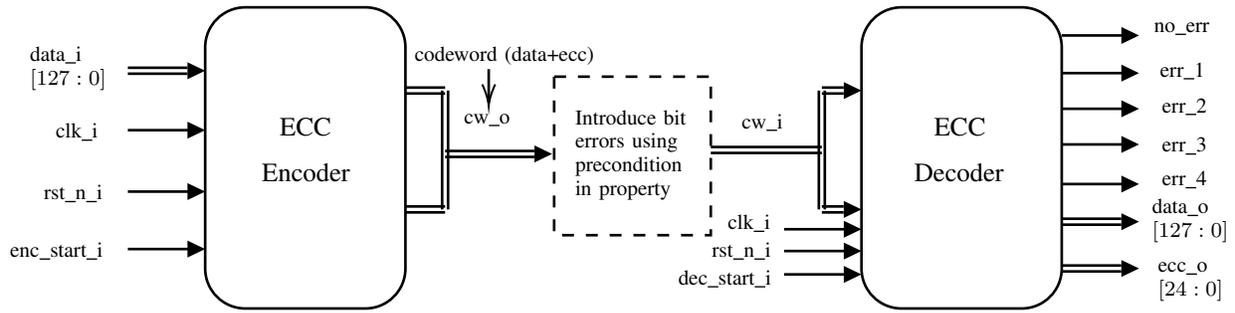
\begin{figure}[h!]
\tikzset{every picture/.style={line width=1pt}} 
\centering
\begin{tikzpicture}[x=0.75pt,y=0.75pt,yscale=-1,xscale=1]

\draw   (110,148.2) .. controls (110,137.04) and (119.04,128) .. (130.2,128) -- (190.8,128) .. controls (201.96,128) and (211,137.04) .. (211,148.2) -- (211,259.8) .. controls (211,270.96) and (201.96,280) .. (190.8,280) -- (130.2,280) .. controls (119.04,280) and (110,270.96) .. (110,259.8) -- cycle ;
\draw   (441,148.2) .. controls (441,137.04) and (450.04,128) .. (461.2,128) -- (521.8,128) .. controls (532.96,128) and (542,137.04) .. (542,148.2) -- (542,259.8) .. controls (542,270.96) and (532.96,280) .. (521.8,280) -- (461.2,280) .. controls (450.04,280) and (441,270.96) .. (441,259.8) -- cycle ;
\draw  [dash pattern={on 4.5pt off 4.5pt}] (286,163) -- (365,163) -- (365,243) -- (286,243) -- cycle ;
\draw    (231,200.5) -- (276,200.5)(231,203.5) -- (276,203.5) ;
\draw [shift={(285,202)}, rotate = 180] [fill={rgb, 255:red, 0; green, 0; blue, 0 }  ][line width=0.08]  [draw opacity=0] (8.93,-4.29) -- (0,0) -- (8.93,4.29) -- cycle    ;
\draw    (232.5,170) -- (232.5,230)(229.5,170) -- (229.5,230) ;
\draw    (211,168.5) -- (231,168.5)(211,171.5) -- (231,171.5) ;
\draw    (211,228.5) -- (231,228.5)(211,231.5) -- (231,231.5) ;
\draw    (421,168.5) -- (432,168.5)(421,171.5) -- (432,171.5) ;
\draw [shift={(441,170)}, rotate = 180] [fill={rgb, 255:red, 0; green, 0; blue, 0 }  ][line width=0.08]  [draw opacity=0] (8.93,-4.29) -- (0,0) -- (8.93,4.29) -- cycle    ;
\draw    (421,228.5) -- (432,228.5)(421,231.5) -- (432,231.5) ;
\draw [shift={(441,230)}, rotate = 180] [fill={rgb, 255:red, 0; green, 0; blue, 0 }  ][line width=0.08]  [draw opacity=0] (8.93,-4.29) -- (0,0) -- (8.93,4.29) -- cycle    ;
\draw    (422.5,170) -- (422.5,230)(419.5,170) -- (419.5,230) ;
\draw    (365,198.5) -- (421,198.5)(365,201.5) -- (421,201.5) ;
\draw    (71,158.5) -- (101,158.5)(71,161.5) -- (101,161.5) ;
\draw [shift={(110,160)}, rotate = 180] [fill={rgb, 255:red, 0; green, 0; blue, 0 }  ][line width=0.08]  [draw opacity=0] (8.93,-4.29) -- (0,0) -- (8.93,4.29) -- cycle    ;
\draw    (71,190) -- (107,190) ;
\draw [shift={(110,190)}, rotate = 180] [fill={rgb, 255:red, 0; green, 0; blue, 0 }  ][line width=0.08]  [draw opacity=0] (8.93,-4.29) -- (0,0) -- (8.93,4.29) -- cycle    ;
\draw    (71,221) -- (107,221) ;
\draw [shift={(110,221)}, rotate = 180] [fill={rgb, 255:red, 0; green, 0; blue, 0 }  ][line width=0.08]  [draw opacity=0] (8.93,-4.29) -- (0,0) -- (8.93,4.29) -- cycle    ;
\draw    (71,250) -- (107,250) ;
\draw [shift={(110,250)}, rotate = 180] [fill={rgb, 255:red, 0; green, 0; blue, 0 }  ][line width=0.08]  [draw opacity=0] (8.93,-4.29) -- (0,0) -- (8.93,4.29) -- cycle    ;
\draw    (402,240) -- (438,240) ;
\draw [shift={(441,240)}, rotate = 180] [fill={rgb, 255:red, 0; green, 0; blue, 0 }  ][line width=0.08]  [draw opacity=0] (8.93,-4.29) -- (0,0) -- (8.93,4.29) -- cycle    ;
\draw    (402,262.8) -- (438,262.8) ;
\draw [shift={(441,262.8)}, rotate = 180] [fill={rgb, 255:red, 0; green, 0; blue, 0 }  ][line width=0.08]  [draw opacity=0] (8.93,-4.29) -- (0,0) -- (8.93,4.29) -- cycle    ;
\draw    (402,251) -- (438,251) ;
\draw [shift={(441,251)}, rotate = 180] [fill={rgb, 255:red, 0; green, 0; blue, 0 }  ][line width=0.08]  [draw opacity=0] (8.93,-4.29) -- (0,0) -- (8.93,4.29) -- cycle    ;
\draw    (542,142.2) -- (578,142.2) ;
\draw [shift={(581,142.2)}, rotate = 180] [fill={rgb, 255:red, 0; green, 0; blue, 0 }  ][line width=0.08]  [draw opacity=0] (8.93,-4.29) -- (0,0) -- (8.93,4.29) -- cycle    ;
\draw    (542,161.2) -- (578,161.2) ;
\draw [shift={(581,161.2)}, rotate = 180] [fill={rgb, 255:red, 0; green, 0; blue, 0 }  ][line width=0.08]  [draw opacity=0] (8.93,-4.29) -- (0,0) -- (8.93,4.29) -- cycle    ;
\draw    (542,179.2) -- (578,179.2) ;
\draw [shift={(581,179.2)}, rotate = 180] [fill={rgb, 255:red, 0; green, 0; blue, 0 }  ][line width=0.08]  [draw opacity=0] (8.93,-4.29) -- (0,0) -- (8.93,4.29) -- cycle    ;
\draw    (542,197.2) -- (578,197.2) ;
\draw [shift={(581,197.2)}, rotate = 180] [fill={rgb, 255:red, 0; green, 0; blue, 0 }  ][line width=0.08]  [draw opacity=0] (8.93,-4.29) -- (0,0) -- (8.93,4.29) -- cycle    ;
\draw    (542,217.2) -- (578,217.2) ;
\draw [shift={(581,217.2)}, rotate = 180] [fill={rgb, 255:red, 0; green, 0; blue, 0 }  ][line width=0.08]  [draw opacity=0] (8.93,-4.29) -- (0,0) -- (8.93,4.29) -- cycle    ;
\draw    (542,234.7) -- (572,234.7)(542,237.7) -- (572,237.7) ;
\draw [shift={(581,236.2)}, rotate = 180] [fill={rgb, 255:red, 0; green, 0; blue, 0 }  ][line width=0.08]  [draw opacity=0] (8.93,-4.29) -- (0,0) -- (8.93,4.29) -- cycle    ;
\draw    (542,258.3) -- (572,258.3)(542,261.3) -- (572,261.3) ;
\draw [shift={(581,259.8)}, rotate = 180] [fill={rgb, 255:red, 0; green, 0; blue, 0 }  ][line width=0.08]  [draw opacity=0] (8.93,-4.29) -- (0,0) -- (8.93,4.29) -- cycle    ;
\draw    (253,159.2) -- (253,175.8) ;
\draw [shift={(253,177.8)}, rotate = 270] [color={rgb, 255:red, 0; green, 0; blue, 0 }  ][line width=0.75]    (10.93,-3.29) .. controls (6.95,-1.4) and (3.31,-0.3) .. (0,0) .. controls (3.31,0.3) and (6.95,1.4) .. (10.93,3.29)   ;

\draw (146,182) node [anchor=north west][inner sep=0.75pt]   [align=left] {ECC};
\draw (137,205) node [anchor=north west][inner sep=0.75pt]   [align=left] {Encoder};
\draw (476,182) node [anchor=north west][inner sep=0.75pt]   [align=left] {ECC};
\draw (466,205) node [anchor=north west][inner sep=0.75pt]   [align=left] {Decoder};
\draw (295,178) node [anchor=north west][inner sep=0.75pt]  [font=\footnotesize] [align=left] {Introduce bit\\errors using\\precondition\\in property};
\draw (20,146) node [anchor=north west][inner sep=0.75pt]  [font=\footnotesize] [align=left] {data\_i\\$[127:0]$};
\draw (33,185) node [anchor=north west][inner sep=0.75pt]  [font=\footnotesize] [align=left] {clk\_i};
\draw (27,216) node [anchor=north west][inner sep=0.75pt]  [font=\footnotesize] [align=left] {rst\_n\_i};
\draw (10,245) node [anchor=north west][inner sep=0.75pt]  [font=\footnotesize] [align=left] {enc\_start\_i};
\draw (372,231) node [anchor=north west][inner sep=0.75pt]  [font=\footnotesize] [align=left] {clk\_i};
\draw (364,245) node [anchor=north west][inner sep=0.75pt]  [font=\footnotesize] [align=left] {rst\_n\_i};
\draw (347,259) node [anchor=north west][inner sep=0.75pt]  [font=\footnotesize] [align=left] {dec\_start\_i};
\draw (587,134) node [anchor=north west][inner sep=0.75pt]  [font=\footnotesize] [align=left] {no\_err};
\draw (590,154) node [anchor=north west][inner sep=0.75pt]  [font=\footnotesize] [align=left] {err\_1};
\draw (591,172) node [anchor=north west][inner sep=0.75pt]  [font=\footnotesize] [align=left] {err\_2};
\draw (591,192) node [anchor=north west][inner sep=0.75pt]  [font=\footnotesize] [align=left] {err\_3};
\draw (591,209) node [anchor=north west][inner sep=0.75pt]  [font=\footnotesize] [align=left] {err\_4};
\draw (586,223) node [anchor=north west][inner sep=0.75pt]  [font=\footnotesize] [align=left] {data\_o\\$[127:0]$};
\draw (589,255) node [anchor=north west][inner sep=0.75pt]  [font=\footnotesize] [align=left] {ecc\_o\\$[24:0]$};
\draw (214,146.2) node [anchor=north west][inner sep=0.75pt]  [font=\footnotesize] [align=left] {codeword (data+ecc)};
\draw (239,180) node [anchor=north west][inner sep=0.75pt]  [font=\footnotesize] [align=left] {cw\_o};
\draw (379,180) node [anchor=north west][inner sep=0.75pt]  [font=\footnotesize] [align=left] {cw\_i};

\end{tikzpicture}
\caption{Sequential ECC core}
\label{seq_ecc}
\end{figure}

A sequential version of the \acrshort{ECC} core is also verified using the formal approach as shown in the Fig.~\ref{seq_ecc}. The core has a sequentially pipelined encoding and decoding stages.

\begin{figure}[h!]
    \centering
    \includegraphics[width=0.4\columnwidth]{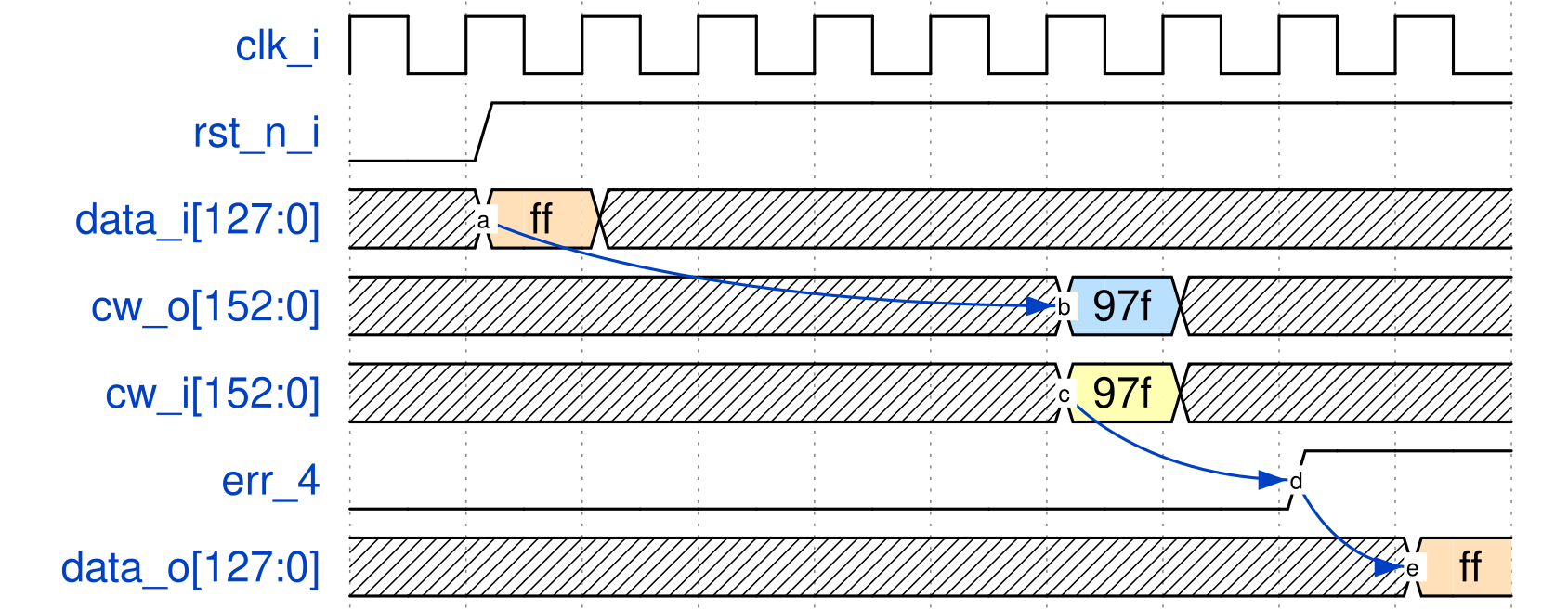}
    \caption{Timing diagram of sequential ECC core}
    \label{timing_diagram}
\end{figure}

Fig.~\ref{timing_diagram} shows the timing diagram of the sequential \acrshort{ECC} core having 6 clock cycles for encoding, 3 clock cycles for error detection and 2 additional clock cycles for error correction. Upon applying the linearity approach, the formal tool gave up after 120 hours with a bounded proof result.

\begin{figure}[h!]
    \centering
    \includegraphics[width=0.4\textwidth]{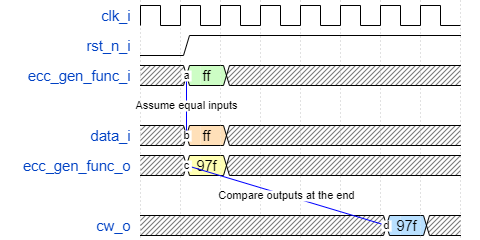}
    \caption{Comparing model against the RTL}
    \label{eq_check}
\end{figure}

It is evident that the encoder has a longer latency in terms of the clock cycles. This in turn increases the state-space for the formal tool and requires more proof time. To overcome this challenge, a cycle inaccurate model of the \acrshort{ECC} encoder is prepared in a combinatorial SystemVerilog function. The model calculates the the \acrshort{ECC} bits in the same clock cycle based on the Reed-Solomon code (as used in the design). In order to prove the correctness of the model, an equivalence check between the combinatorial function and the sequential encoding output from the \acrshort{RTL} design is performed by assuming equal inputs for both and comparing the outputs at the end as shown in Fig.~\ref{eq_check}. The \acrshort{SVA} property to prove the equivalence is mentioned in the Listing~\ref{equivalence}.

\vspace{0.25cm}
\begin{lstlisting}[language=Verilog, float=h!, caption=\acrshort{SVA} property to prove the equivalence between RTL and the cycle inaccurate model, basicstyle=\ttfamily, label={equivalence}]
//=================================================
// Prove equivalence between RTL and model
//=================================================
property equivalence;
    (data_i == 128'hcaba79a638aca501) 
  |->
    ##5 (({ecc_gen_func(data_i), data_i}) == cw_o);
endproperty
ap_equivalence : assert property(@(posedge clk) disable iff (!rst_n) equivalence);
\end{lstlisting}

The model is used afterwards in the properties as a glue logic to prove the correctness of the design.

\subsection{\textbf{Addressing Complexity Step 4: Use SST Traces to Help with Induction-Based Proof}}
For sequentially deep designs such as the sequential \acrshort{ECC} used in this work, it might get difficult to get an unbounded proof result within reasonable time. In spite of using the above mentioned complexity reduction techniques, a few properties such as the 4 bit error correction struggles to prove. To overcome this problem, an induction-based proof method or k-induction is used. Induction is a method to check if the design is in a random good state whether it will be in a good state at the next cycle \cite{cadence}. k-induction bounded model checking consists of two steps. These are the base case and the induction step \cite{induction_paper2}. To begin, rather than just one state, the property is assumed to hold for a path of $n$ successive states. This means that a more extensive base-case must be demonstrated. Second, the path's states are assumed to be distinct. Finiteness implies that the second strengthening completes the method in the sense that there is always a length for which the induction-step is provable \cite{induction_paper1}. This can be formalized as equations (\ref{base}) and (\ref{step}) \cite{induction_paper1}:

\begin{equation}
\text {Base}_n \quad:=\mathbf{I}_0 \wedge\left(\left(\mathbf{P}_0 \wedge \mathbf{T}_0\right) \wedge \ldots \wedge\left(\mathbf{P}_{n-1} \wedge \mathbf{T}_{n-1}\right)\right) \wedge \overline{\mathbf{P}_n}
\label{base}
\end{equation}

\begin{equation}
\text {Step}_n \quad:=\left(\left(\mathbf{P}_0 \wedge \mathbf{T}_0\right) \wedge \ldots \wedge\left(\mathbf{P}_n \wedge \mathbf{T}_n\right)\right) \wedge \overline{\mathbf{P}_{n+1}}
\label{step}
\end{equation}

where ${I}_0$ is the initial state, ${P}_0$ is the property in initial state and ${T}_0$ is time zero. $n$ is the time step. The interpretation of these formulas/equations is mentioned in Fig.~\ref{formula}. If the $n$-th base-case is unsatisfiable, the statement should be interpreted as \say{There exists no $n$-step path to a state violating the property, assuming the property holds the first $n$-1 steps}. If the $n$-th induction-step is unsatisfiable, it should be read as \say{Following an $n$-step trace where the property holds, there exists no next state where it fails}. SAT solvers are used in modern \acrshort{EDA} tools to solve such induction equations.

\tikzset{every picture/.style={line width=1pt}} 
\begin{figure}[h!]
\centering
\begin{tikzpicture}[x=0.75pt,y=0.75pt,yscale=-1,xscale=1]

\draw   (132,123) .. controls (132,118.03) and (136.03,114) .. (141,114) .. controls (145.97,114) and (150,118.03) .. (150,123) .. controls (150,127.97) and (145.97,132) .. (141,132) .. controls (136.03,132) and (132,127.97) .. (132,123) -- cycle ;
\draw   (170,123) .. controls (170,118.03) and (174.03,114) .. (179,114) .. controls (183.97,114) and (188,118.03) .. (188,123) .. controls (188,127.97) and (183.97,132) .. (179,132) .. controls (174.03,132) and (170,127.97) .. (170,123) -- cycle ;
\draw    (141,114) .. controls (156.68,83.6) and (170.09,101.94) .. (177.83,112.42) ;
\draw [shift={(179,114)}, rotate = 233.13] [color={rgb, 255:red, 0; green, 0; blue, 0 }  ][line width=0.75]    (10.93,-3.29) .. controls (6.95,-1.4) and (3.31,-0.3) .. (0,0) .. controls (3.31,0.3) and (6.95,1.4) .. (10.93,3.29)   ;
\draw   (208,123) .. controls (208,118.03) and (212.03,114) .. (217,114) .. controls (221.97,114) and (226,118.03) .. (226,123) .. controls (226,127.97) and (221.97,132) .. (217,132) .. controls (212.03,132) and (208,127.97) .. (208,123) -- cycle ;
\draw    (179,114) .. controls (194.68,83.6) and (208.09,101.94) .. (215.83,112.42) ;
\draw [shift={(217,114)}, rotate = 233.13] [color={rgb, 255:red, 0; green, 0; blue, 0 }  ][line width=0.75]    (10.93,-3.29) .. controls (6.95,-1.4) and (3.31,-0.3) .. (0,0) .. controls (3.31,0.3) and (6.95,1.4) .. (10.93,3.29)   ;
\draw    (217,114) .. controls (232.68,83.6) and (246.09,101.94) .. (253.83,112.42) ;
\draw [shift={(255,114)}, rotate = 233.13] [color={rgb, 255:red, 0; green, 0; blue, 0 }  ][line width=0.75]    (10.93,-3.29) .. controls (6.95,-1.4) and (3.31,-0.3) .. (0,0) .. controls (3.31,0.3) and (6.95,1.4) .. (10.93,3.29)   ;
\draw   (320,123) .. controls (320,118.03) and (324.03,114) .. (329,114) .. controls (333.97,114) and (338,118.03) .. (338,123) .. controls (338,127.97) and (333.97,132) .. (329,132) .. controls (324.03,132) and (320,127.97) .. (320,123) -- cycle ;
\draw    (291,114) .. controls (306.68,83.6) and (320.09,101.94) .. (327.83,112.42) ;
\draw [shift={(329,114)}, rotate = 233.13] [color={rgb, 255:red, 0; green, 0; blue, 0 }  ][line width=0.75]    (10.93,-3.29) .. controls (6.95,-1.4) and (3.31,-0.3) .. (0,0) .. controls (3.31,0.3) and (6.95,1.4) .. (10.93,3.29)   ;
\draw   (358,123) .. controls (358,118.03) and (362.03,114) .. (367,114) .. controls (371.97,114) and (376,118.03) .. (376,123) .. controls (376,127.97) and (371.97,132) .. (367,132) .. controls (362.03,132) and (358,127.97) .. (358,123) -- cycle ;
\draw    (329,114) .. controls (344.68,83.6) and (358.09,101.94) .. (365.83,112.42) ;
\draw [shift={(367,114)}, rotate = 233.13] [color={rgb, 255:red, 0; green, 0; blue, 0 }  ][line width=0.75]    (10.93,-3.29) .. controls (6.95,-1.4) and (3.31,-0.3) .. (0,0) .. controls (3.31,0.3) and (6.95,1.4) .. (10.93,3.29)   ;
\draw   (132,243) .. controls (132,238.03) and (136.03,234) .. (141,234) .. controls (145.97,234) and (150,238.03) .. (150,243) .. controls (150,247.97) and (145.97,252) .. (141,252) .. controls (136.03,252) and (132,247.97) .. (132,243) -- cycle ;
\draw   (170,243) .. controls (170,238.03) and (174.03,234) .. (179,234) .. controls (183.97,234) and (188,238.03) .. (188,243) .. controls (188,247.97) and (183.97,252) .. (179,252) .. controls (174.03,252) and (170,247.97) .. (170,243) -- cycle ;
\draw    (141,234) .. controls (156.68,203.6) and (170.09,221.94) .. (177.83,232.42) ;
\draw [shift={(179,234)}, rotate = 233.13] [color={rgb, 255:red, 0; green, 0; blue, 0 }  ][line width=0.75]    (10.93,-3.29) .. controls (6.95,-1.4) and (3.31,-0.3) .. (0,0) .. controls (3.31,0.3) and (6.95,1.4) .. (10.93,3.29)   ;
\draw   (208,243) .. controls (208,238.03) and (212.03,234) .. (217,234) .. controls (221.97,234) and (226,238.03) .. (226,243) .. controls (226,247.97) and (221.97,252) .. (217,252) .. controls (212.03,252) and (208,247.97) .. (208,243) -- cycle ;
\draw    (179,234) .. controls (194.68,203.6) and (208.09,221.94) .. (215.83,232.42) ;
\draw [shift={(217,234)}, rotate = 233.13] [color={rgb, 255:red, 0; green, 0; blue, 0 }  ][line width=0.75]    (10.93,-3.29) .. controls (6.95,-1.4) and (3.31,-0.3) .. (0,0) .. controls (3.31,0.3) and (6.95,1.4) .. (10.93,3.29)   ;
\draw    (217,234) .. controls (232.68,203.6) and (246.09,221.94) .. (253.83,232.42) ;
\draw [shift={(255,234)}, rotate = 233.13] [color={rgb, 255:red, 0; green, 0; blue, 0 }  ][line width=0.75]    (10.93,-3.29) .. controls (6.95,-1.4) and (3.31,-0.3) .. (0,0) .. controls (3.31,0.3) and (6.95,1.4) .. (10.93,3.29)   ;
\draw   (320,243) .. controls (320,238.03) and (324.03,234) .. (329,234) .. controls (333.97,234) and (338,238.03) .. (338,243) .. controls (338,247.97) and (333.97,252) .. (329,252) .. controls (324.03,252) and (320,247.97) .. (320,243) -- cycle ;
\draw    (291,234) .. controls (306.68,203.6) and (320.09,221.94) .. (327.83,232.42) ;
\draw [shift={(329,234)}, rotate = 233.13] [color={rgb, 255:red, 0; green, 0; blue, 0 }  ][line width=0.75]    (10.93,-3.29) .. controls (6.95,-1.4) and (3.31,-0.3) .. (0,0) .. controls (3.31,0.3) and (6.95,1.4) .. (10.93,3.29)   ;
\draw   (358,243) .. controls (358,238.03) and (362.03,234) .. (367,234) .. controls (371.97,234) and (376,238.03) .. (376,243) .. controls (376,247.97) and (371.97,252) .. (367,252) .. controls (362.03,252) and (358,247.97) .. (358,243) -- cycle ;
\draw    (329,234) .. controls (344.68,203.6) and (358.09,221.94) .. (365.83,232.42) ;
\draw [shift={(367,234)}, rotate = 233.13] [color={rgb, 255:red, 0; green, 0; blue, 0 }  ][line width=0.75]    (10.93,-3.29) .. controls (6.95,-1.4) and (3.31,-0.3) .. (0,0) .. controls (3.31,0.3) and (6.95,1.4) .. (10.93,3.29)   ;
\draw   (396,243) .. controls (396,238.03) and (400.03,234) .. (405,234) .. controls (409.97,234) and (414,238.03) .. (414,243) .. controls (414,247.97) and (409.97,252) .. (405,252) .. controls (400.03,252) and (396,247.97) .. (396,243) -- cycle ;
\draw    (367,234) .. controls (382.68,203.6) and (396.09,221.94) .. (403.83,232.42) ;
\draw [shift={(405,234)}, rotate = 233.13] [color={rgb, 255:red, 0; green, 0; blue, 0 }  ][line width=0.75]    (10.93,-3.29) .. controls (6.95,-1.4) and (3.31,-0.3) .. (0,0) .. controls (3.31,0.3) and (6.95,1.4) .. (10.93,3.29)   ;
\draw    (356.5,146) -- (367.5,146) ;
\draw    (367.5,146) -- (367.5,151) ;
\draw    (394.5,266) -- (405.5,266) ;
\draw    (405.5,266) -- (405.5,271) ;
\draw  [dash pattern={on 0.84pt off 2.51pt}]  (264,120) -- (285.5,120) ;
\draw  [dash pattern={on 0.84pt off 2.51pt}]  (264,242) -- (285.5,242) ;

\draw (130,140) node [anchor=north west][inner sep=0.75pt]   [align=left] {I, P};
\draw (173,140) node [anchor=north west][inner sep=0.75pt]   [align=left] {P};
\draw (211,140) node [anchor=north west][inner sep=0.75pt]   [align=left] {P};
\draw (323,140) node [anchor=north west][inner sep=0.75pt]   [align=left] {P};
\draw (370,140) node [anchor=north west][inner sep=0.75pt]   [align=left] {P};
\draw (130,260) node [anchor=north west][inner sep=0.75pt]   [align=left] {I, P};
\draw (173,260) node [anchor=north west][inner sep=0.75pt]   [align=left] {P};
\draw (211,260) node [anchor=north west][inner sep=0.75pt]   [align=left] {P};
\draw (323,260) node [anchor=north west][inner sep=0.75pt]   [align=left] {P};
\draw (408,260) node [anchor=north west][inner sep=0.75pt]   [align=left] {P};
\draw (361,260) node [anchor=north west][inner sep=0.75pt]   [align=left] {P};
\draw (155,75) node [anchor=north west][inner sep=0.75pt]   [align=left] {T};
\draw (192,75) node [anchor=north west][inner sep=0.75pt]   [align=left] {T};
\draw (230,75) node [anchor=north west][inner sep=0.75pt]   [align=left] {T};
\draw (303,75) node [anchor=north west][inner sep=0.75pt]   [align=left] {T};
\draw (342,75) node [anchor=north west][inner sep=0.75pt]   [align=left] {T};
\draw (154,195) node [anchor=north west][inner sep=0.75pt]   [align=left] {T};
\draw (191,195) node [anchor=north west][inner sep=0.75pt]   [align=left] {T};
\draw (229,195) node [anchor=north west][inner sep=0.75pt]   [align=left] {T};
\draw (302,195) node [anchor=north west][inner sep=0.75pt]   [align=left] {T};
\draw (341,195) node [anchor=north west][inner sep=0.75pt]   [align=left] {T};
\draw (380,196) node [anchor=north west][inner sep=0.75pt]   [align=left] {T};
\draw (131,171) node [anchor=north west][inner sep=0.75pt]   [align=left] {\textbf{Induction-step}};
\draw (131,53) node [anchor=north west][inner sep=0.75pt]   [align=left] {\textbf{Base-case}};

\end{tikzpicture}
\caption{Base-case and induction-step \cite{induction_paper1}}
\label{formula}
\end{figure}
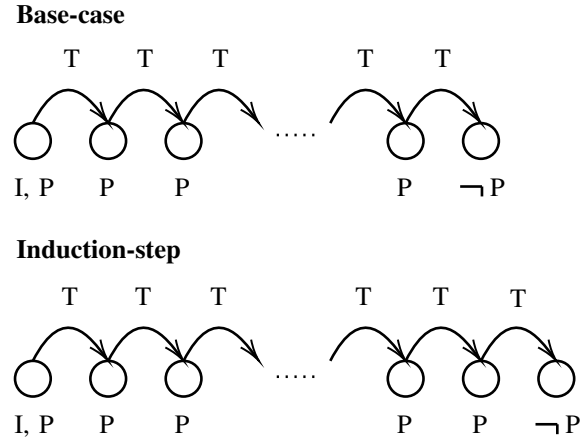

In a simple and abstract explanation to understand proof using induction, assume the equations (\ref{stepa}) and (\ref{stepb}) about a design:

\begin{equation}
A_1=1
\label{stepa}
\end{equation}

\begin{equation}
\left(A_n \rightarrow A_{n+1}\right)=1
\label{stepb}
\end{equation}

Step (\ref{stepa}) is proving whether the assertion holds true for the 1st clock cycle and step (\ref{stepb}) checks to prove whether the assertion $A$ holds true on a random state $A_n$ which would mean it will hold true for the next clock cycle ($A_{n+1}$) as well. If step (\ref{stepa}) and (\ref{stepb}) are true, this proves that $A$ is true for all clock cycles \cite{cadence}.

Modern \acrshort{EDA} tools such as Cadence JasperGold offer solutions like the \acrfull{SST} based on induction-based proof where a systematic process is used to reduce the state space of a target property. The reduction is essentially done by identifying such states and writing helper assertions called as lemmas. Helper assertions are properties about internal signals (states) in the design that, if proven, will be used as \say{assumptions} in the proof process of the target property (or other properties), allowing these proven properties to aid in the elimination of states \cite{cadence}. An \acrshort{SST} trace is a normal \acrfull{CEX} of the property, but since it does not start at reset states, this trace exemplifies the target property failure \cite{cadence}:

\begin{itemize}
    \item For a true property, this trace starts from an unreachable good state and ends outside the property in a bad state.
    \item For a false-property, the trace can start from reachable good state and end in reachable bad state.
\end{itemize}

As the \acrshort{SST} process normally focuses on true properties, the trace is a normal \acrshort{CEX} that starts at an unreachable state, continues for $N$-1 cycles in unreachable states, then violates the target property on the last cycle. These \acrshort{SST} traces helps to write the helper assertions. Additionally, the \acrshort{SST} process provides a deeper understanding of the design and relationships that exist between its state variables. This deeper understanding can help the user tune the model and/or prove other properties beyond the target property \cite{cadence}. A usual \acrshort{SST} based flow is depicted in Fig.~\ref{k_induction}.

\tikzset{every picture/.style={line width=1pt}} 
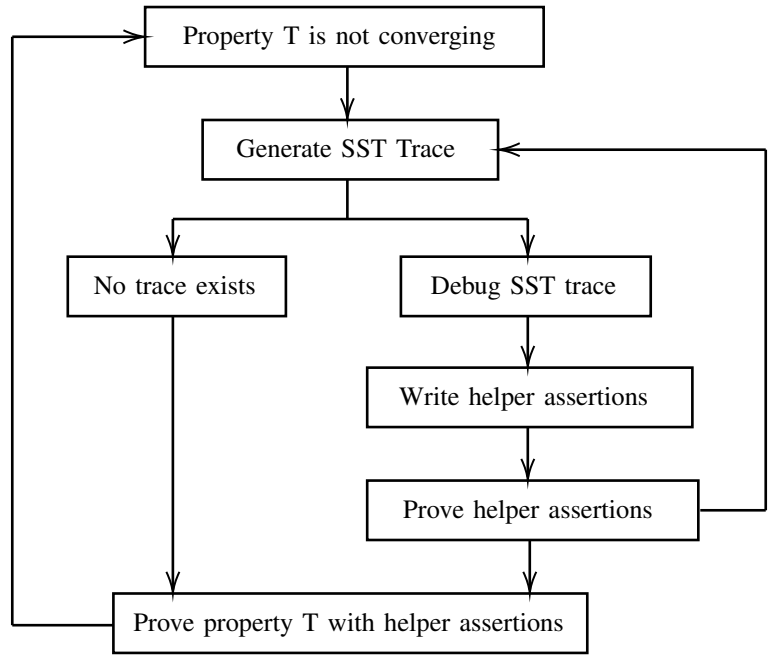
\begin{figure}[h!]
\centering
\begin{tikzpicture}[x=0.75pt,y=0.75pt,yscale=-1,xscale=1]

\draw   (217.5,24) -- (418.5,24) -- (418.5,54) -- (217.5,54) -- cycle ;
\draw   (246.5,81) -- (395.5,81) -- (395.5,111) -- (246.5,111) -- cycle ;
\draw   (178.5,150) -- (288.5,150) -- (288.5,180) -- (178.5,180) -- cycle ;
\draw   (346.5,150) -- (472.5,150) -- (472.5,180) -- (346.5,180) -- cycle ;
\draw   (329.5,206) -- (493.5,206) -- (493.5,236) -- (329.5,236) -- cycle ;
\draw   (329.5,263) -- (496.5,263) -- (496.5,293) -- (329.5,293) -- cycle ;
\draw   (201.5,320) -- (440.5,320) -- (440.5,350) -- (201.5,350) -- cycle ;
\draw    (319.5,54) -- (319.5,79) ;
\draw [shift={(319.5,81)}, rotate = 270] [color={rgb, 255:red, 0; green, 0; blue, 0 }  ][line width=0.75]    (10.93,-3.29) .. controls (6.95,-1.4) and (3.31,-0.3) .. (0,0) .. controls (3.31,0.3) and (6.95,1.4) .. (10.93,3.29)   ;
\draw    (410.5,180) -- (410.5,204) ;
\draw [shift={(410.5,206)}, rotate = 270] [color={rgb, 255:red, 0; green, 0; blue, 0 }  ][line width=0.75]    (10.93,-3.29) .. controls (6.95,-1.4) and (3.31,-0.3) .. (0,0) .. controls (3.31,0.3) and (6.95,1.4) .. (10.93,3.29)   ;
\draw    (410.5,236) -- (410.5,261) ;
\draw [shift={(410.5,263)}, rotate = 270] [color={rgb, 255:red, 0; green, 0; blue, 0 }  ][line width=0.75]    (10.93,-3.29) .. controls (6.95,-1.4) and (3.31,-0.3) .. (0,0) .. controls (3.31,0.3) and (6.95,1.4) .. (10.93,3.29)   ;
\draw    (411.5,293) -- (411.5,318) ;
\draw [shift={(411.5,320)}, rotate = 270] [color={rgb, 255:red, 0; green, 0; blue, 0 }  ][line width=0.75]    (10.93,-3.29) .. controls (6.95,-1.4) and (3.31,-0.3) .. (0,0) .. controls (3.31,0.3) and (6.95,1.4) .. (10.93,3.29)   ;
\draw    (231.5,180) -- (231.5,318) ;
\draw [shift={(231.5,320)}, rotate = 270] [color={rgb, 255:red, 0; green, 0; blue, 0 }  ][line width=0.75]    (10.93,-3.29) .. controls (6.95,-1.4) and (3.31,-0.3) .. (0,0) .. controls (3.31,0.3) and (6.95,1.4) .. (10.93,3.29)   ;
\draw    (150.5,39) -- (150.5,336) ;
\draw    (150.5,39) -- (215.5,39) ;
\draw [shift={(217.5,39)}, rotate = 180] [color={rgb, 255:red, 0; green, 0; blue, 0 }  ][line width=0.75]    (10.93,-3.29) .. controls (6.95,-1.4) and (3.31,-0.3) .. (0,0) .. controls (3.31,0.3) and (6.95,1.4) .. (10.93,3.29)   ;
\draw    (150.5,336) -- (201.5,336) ;
\draw    (497.5,278) -- (530.5,278) ;
\draw    (530.5,96) -- (530.5,278) ;
\draw    (530.5,96) -- (397.5,96) ;
\draw [shift={(395.5,96)}, rotate = 360] [color={rgb, 255:red, 0; green, 0; blue, 0 }  ][line width=0.75]    (10.93,-3.29) .. controls (6.95,-1.4) and (3.31,-0.3) .. (0,0) .. controls (3.31,0.3) and (6.95,1.4) .. (10.93,3.29)   ;
\draw    (231.5,131) -- (410.5,131) ;
\draw    (410.5,131) -- (410.5,148) ;
\draw [shift={(410.5,150)}, rotate = 270] [color={rgb, 255:red, 0; green, 0; blue, 0 }  ][line width=0.75]    (10.93,-3.29) .. controls (6.95,-1.4) and (3.31,-0.3) .. (0,0) .. controls (3.31,0.3) and (6.95,1.4) .. (10.93,3.29)   ;
\draw    (231.5,131) -- (231.5,148) ;
\draw [shift={(231.5,150)}, rotate = 270] [color={rgb, 255:red, 0; green, 0; blue, 0 }  ][line width=0.75]    (10.93,-3.29) .. controls (6.95,-1.4) and (3.31,-0.3) .. (0,0) .. controls (3.31,0.3) and (6.95,1.4) .. (10.93,3.29)   ;
\draw    (319.5,111) -- (319.5,131) ;

\draw (235,32) node [anchor=north west][inner sep=0.75pt]   [align=left] {Property T is not converging};
\draw (262,89) node [anchor=north west][inner sep=0.75pt]   [align=left] {Generate SST Trace};
\draw (190,158) node [anchor=north west][inner sep=0.75pt]   [align=left] {No trace exists};
\draw (360,158) node [anchor=north west][inner sep=0.75pt]   [align=left] {Debug SST trace};
\draw (344,214) node [anchor=north west][inner sep=0.75pt]   [align=left] {Write helper assertions};
\draw (346,271) node [anchor=north west][inner sep=0.75pt]   [align=left] {Prove helper assertions};
\draw (210,328) node [anchor=north west][inner sep=0.75pt]   [align=left] {Prove property T with helper assertions};

\end{tikzpicture}
\caption{k-Induction and \acrshort{SST} based verification flow \cite{cadence}}
\label{k_induction}
\end{figure}

%

\begin{figure}[h!]
    \centering
    \includegraphics[width=0.3\textwidth]{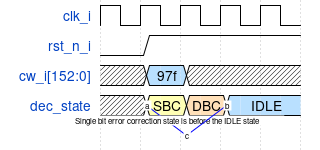}
    \caption{Decoder FSM did not start from IDLE state when decoding starts}
    \label{fsm}
\end{figure}
\begin{figure}[h!]
    \centering
    \includegraphics[width=0.5\textwidth]{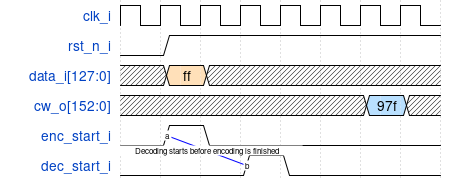}
    \caption{Decoding starts before encoding is finished}
    \label{decoding}
\end{figure}

While analyzing the \acrshort{SST} trace for the \acrshort{ECC} design, two issues were found in the violation due to missing constraints:

\begin{itemize}
  \item The decoder \acrshort{FSM} should start from IDLE state when decoding starts (Fig.~\ref{fsm}).
  \item Decoding should start only after encoding is finished  (Fig.~\ref{decoding}).
\end{itemize}

As a result the induction steps fail for increasingly longer paths, contributing towards longer proof runtime. After adding these two missing constraints in Listings~\ref{dec_idle} and \ref{enc_dec_no_overlap}, all the properties proved within 24 hours.
\vspace{0.25cm}
\begin{lstlisting}[language=Verilog, float=h!, caption=\acrshort{SVA} property to assume that the decoder starts from the IDLE state, basicstyle=\ttfamily, label={dec_idle}]
//===================================================================
// Assume that the decoder is in IDLE state when decoding is started
//===================================================================
property dec_idle_state;
    ( (dec_start_i)
  |->
    ( dec_state == IDLE ));
endproperty
cp_dec_idle_state : assume property(dec_idle_state);
\end{lstlisting}
\begin{lstlisting}[language=Verilog, float=h!, caption=\acrshort{SVA} property to assume that encoding and decoding stages do not overlap, basicstyle=\ttfamily, label={enc_dec_no_overlap}]
//==================================================
// Assume that encoding and decoding do not overlap
//==================================================
property no_overlap;
    ( (enc_start_i)
  |->
    ( !dec_data_valid_i ) [*5] );
endproperty
cp_no_overlap : assume property(@(posedge clk) disable iff (!rst_n) no_overlap);
\end{lstlisting}

\section{Conclusion and Future Work}
ECCs are commonly used safety mechanism to ensure the functional safety. An exhaustive verification of such safety-critical deisgn is very important. A pragmatic formal verification approach for complex \acrshort{ECC} designs is presented in this paper. Several \acrshort{ECC} cores with combinatorial and sequentially pipelined ecoding/decoding stages are verified using the proposed approach. The linearity of the syndrome generator is used to limit the input data to a reasonable analysis space and make the design formal-friendly. An equivalent cycle inaccurate model for sequential ECCs is used to reduce the latency for the formal tool and prove the properties in less proof time. During the course of verification, some design bugs such as incorrect one-hot encoding of the error flags were unveiled and fixed before the tape-out. It may be noted that for different formal tools, selection and tuning of the appropriate engine settings also plays an important role. Since \acrshort{EDA} tools are becoming more advanced and deploy different engines to solve specific mathematical problems related to the properties, a right set of engines also helps in a faster proof time. For the \acrshort{ECC} verified in this paper, it took 24 hours to prove all the properties together with the complexity reduction techniques discussed whereas, it took 18 hours to prove the properties with dedicated engine settings.

Although the proposed verification method is scalable and works on standard \acrshort{ECC} designs, it is noted that the designs having higher data width, or higher number of bits to detect and correct, or higher latency in encoding/decoding stages, or all of them, make the tool suffer to give an unbounded proof result within reasonable time. As a future work, one could look into the possibility of using even better abstract models such as C/C++/SystemC and use software based formal verification tools such as CBMC \cite{cbmc} or ESBMC \cite{esbmc} to prove the properties by exploiting word-level nature of C/C++ and speedup the verification process. Later on, an equivalence check between the \acrshort{RTL} and the abstract model could be done to prove the correctness of the design.

\printbibliography

\end{document}